%% file: main.tex
\documentclass{article}

\usepackage[T1]{fontenc}

\usepackage{iclr2027_conference,times}   %

\usepackage{amsmath}
\usepackage{amssymb}
\usepackage{graphicx}
\usepackage{booktabs}
\usepackage{xcolor}
\usepackage{url}
\usepackage{etoolbox}
\usepackage{adjustbox}
\usepackage[hidelinks]{hyperref}

\BeforeBeginEnvironment{tabular}{\begin{adjustbox}{max width=\linewidth}}
\AfterEndEnvironment{tabular}{\end{adjustbox}}

\newcommand{\prrm}{\mathrm{PRR}_{\mathrm{matched}}}
\newcommand{\prrd}{\mathrm{PRR}_{\mathrm{deployed}}}
\newcommand{\prrocc}{\mathrm{PRR}_{\mathrm{occ}}}
\newcommand{\gsm}{\texttt{gsm\_hard}}
\newcommand{\wi}{\texttt{wrong\_intermediate}}

\title{Re-derivability Decides What a Staged Agent Pipeline Recovers\\
After an Upstream Fault}

\input{authors}

\iclrfinalcopy

\makeatletter
\patchcmd{\@maketitle}{Published as a conference paper at ICLR 2027}{Preprint. Under review.}{}{\errmessage{arXiv preprint header patch failed}}
\makeatother

\begin{document}

\maketitle

\begin{abstract}
One variable sets what an upstream fault costs a staged pipeline of language-model agents:
\emph{re-derivability}, how much of what a stage needs it can rebuild from the original
problem. Grounding an inspector agent in that problem is worth $+0.608$ $[+0.517, +0.700]$ to
$+0.358$ over a blind one on four open-weight backbones served with thinking disabled, and on
the two Qwen backbones the blind
inspector changes no item at all. The inspector helps only when it can see the problem. That head-to-head is exploratory.
We manipulate it directly. One deterministic fault enters the first stage, and we re-expose the
original problem to $k = 0,\dots,3$ of the downstream stages with agents, items, fault and topology
held fixed, on 120 \gsm{} items per arm at temperature zero. Accuracy under fault rises on four of
four backbones, from $+0.233$ to $+0.392$, the largest Holm-adjusted $p$ being
$2.1{\times}10^{-6}$. A registered kill test
rules out tokens. Blanking every word holds the word slots fixed, and retention tracks the visible
fraction on four of four, climbing from 0.221 to 0.692 on the primary. The prompt-token ratio
between the ends runs 0.90 to 1.03, above one only on Phi-4, whose masked arm still scores worse. Those two families are confirmatory and everything else here is exploratory. The interaction
excludes zero on two of four backbones under the registered pipeline, four of four under a
three-stage pipeline, and three of four under full message history, which passes prior stage outputs, the primary at
$+0.317$. On
Llama-3.1-8B the fault carries no detectable cost at any dose, so the other three carry every
claim about what a fault costs.
Re-derivability also sets what the architecture costs, and no decomposition we measured reliably
beats one direct call. With no fault injected the registered pipeline loses to that call by
$-0.267$, $-0.125$ and $-0.317$, and on Phi-4 reads $+0.058$ at $p = 0.118$, which the test fails to separate from zero. The repair that works
is cheap and front-loaded: the first re-grounded stage buys $+0.394$ of matched retention for
$+59.8$ tokens per item on Qwen3-14B, and the stages after it buy nothing.
\end{abstract}

\section{Introduction}

Teams split a task across staged language-model agents and ship the pipeline. Splitting it
strictly costs accuracy before anything breaks: on 120 \gsm{} items with no fault injected the
pipeline loses to a single direct call on three of four backbones, and on the fourth the
difference is not separable from zero --- a contrast Section~\ref{sec:decompositions} gives in
full, with every interval (Table~\ref{tab:strict-direct}). A fault makes it worse. In a \emph{strict} staged pipeline, where
each stage after the first sees only its predecessor's output, one deterministic fault at the
first stage is expensive: retention against the matched no-fault cell at the same depth is 0.368,
0.289, 0.230 and 0.667 on Qwen3-14B, Qwen3-8B, Phi-4 and Llama-3.1-8B-Instruct
(Table~\ref{tab:headline}, Figure~\ref{fig:hero} in the appendix).

\begin{figure}[t]
\centering
\includegraphics[width=1.0\textwidth]{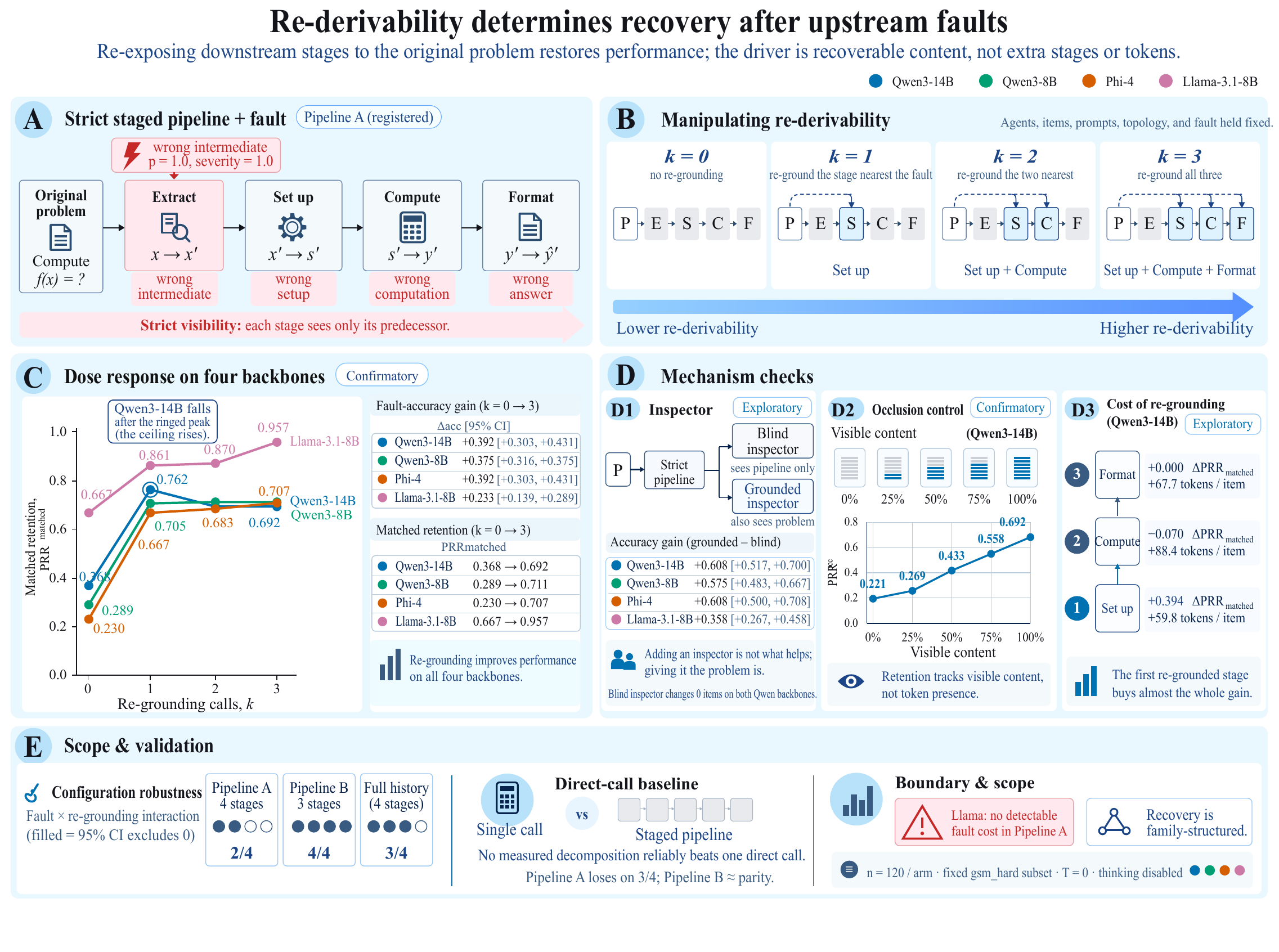}
\caption{Re-grounding a faulted pipeline; each panel's title names what it draws. Every dose in
(B) also runs without the fault at the same depth and with the same re-grounded stages, and that
cell is the ceiling in (C). \textbf{(C)} and \textbf{(D2)} are the two registered confirmatory
families; \textbf{(D1)}, \textbf{(D3)} and \textbf{(E)} are exploratory, (D3) on the primary
backbone alone. Each fraction in (E) counts backbones out of four under the configuration named
beside it. Every arm runs the same fixed 120-item \gsm{} subset
(Section~\ref{sec:setup}).}
\label{fig:mech}
\end{figure}

Trace-corpus taxonomies and attribution tools catalogue such failures
\citep{cemri2025why,albayaydh2026beyond,bisconti2025taxonomy,zhang2025agentracer,qiao2026verifymas};
three lines of work try to explain them and none varies what a stage can see: a closed-loop
advantage explained by a shared message pool \citep{jia2026masfire}, architectural vulnerability
argued as a property of the graph \citep{bappy2026adversarial}, and a grounded inspector measured
without varying its access \citep{huang2025resilience}. Which property decides whether an upstream
mistake survives is open.

\emph{Re-derivability}, how much of what a stage needs it can rebuild from the original problem,
sets what the fault costs. Strict visibility strips it: a stage sees only its predecessor's
message, so a faulty message is all it has. \emph{Re-grounding} raises it. A switch re-injects the
original problem into $k$ of the downstream stages, $k = 0,\dots,3$, with everything else held
fixed and each dose carrying its own
matched no-fault cell at the same depth. At $k = 3$ retention is 0.692, 0.711, 0.707
and 0.957, and the dose-response holds on four of four backbones with the endpoint contrast
rejecting at an adjusted $p$ no larger than $2.1{\times}10^{-6}$ (Figure~\ref{fig:mech}A--C).

The recovery is not a token-count artefact. Replacing every word of the problem by a blank leaves
the same number of word slots and no content; retention there falls to 0.221 on the primary and
rises with the visible fraction on every backbone (Figure~\ref{fig:controls}, left panel;
Section~\ref{sec:kill}). The registered kill test for that risk did not fire.

\paragraph{Contributions.}
\begin{enumerate}\itemsep0pt\parskip0pt\topsep2pt\partopsep0pt
\item \emph{Exploratory.} The strict pipeline loses to one direct call on three of four backbones
  with no fault present, and never reliably wins.
\item \emph{Confirmatory.} Re-derivability sets what a fault costs: a dose-graded manipulation
  with a matched no-fault ceiling at every dose, four of four.
\item \emph{Confirmatory.} The price is re-derivability and not tokens. The registered kill test
  holds token presence fixed and varies only content, passing on four of four.
\item \emph{Exploratory.} A repair must hand back the \emph{problem}. Full history passes the
  transcript instead and carries the fault with it, and robustness-aware topology search ties
  accuracy-only search ($+0.0$ points) and loses to random selection ($-0.6$ points).
\item \emph{Exploratory.} The repair that works is the single call in disguise. Re-grounding
  buys $+0.394$ of matched retention on the primary for $+59.8$ tokens per item, and the stages
  after it buy nothing.
\end{enumerate}

\section{Setup}
\label{sec:setup}

\paragraph{The pipelines and the fault.}
Pipeline~A, the registered one, runs four stages in sequence (extract, set up, compute, format),
each receiving only its predecessor's output. Pipeline~B is a three-stage decomposition of the
same task (understand, solve, answer) under the same visibility rule, fault and items, measured on
all four backbones and reported in Section~\ref{sec:pipelines}. The re-grounding switch re-injects
the original problem into the $k$ downstream stages \emph{nearest the fault}, with
$k_{\max} = 3$ on A and $2$ on B, so $k = 1$ re-grounds the stage that receives the faulted
message directly. The
re-grounded prompt keeps the system line saying the stage does not have the original problem
beside a user line offering it for reference; the scaffold carries that contradiction in every
arm, and we disclose it rather than repairing it mid-campaign. We inject one
fault, \wi{}, at the first stage at probability 1.0 and severity 1.0: a bogus intermediate
step plus a forced answer line carrying a wrong value. It is weaker on families whose upstream
text carries no number, and we claim no single fault across families
(Section~\ref{sec:flat}).

\paragraph{Three ratios, named apart everywhere.}
Let $\mathrm{acc}(F,k)$ be accuracy under fault with $k$ re-grounded stages and $\mathrm{acc}(C,k)$
the no-fault cell at the same depth. We report three ratios:
$\prrm(k) = \mathrm{acc}(F,k)/\mathrm{acc}(C,k)$,
$\prrd(k) = \mathrm{acc}(F,k)/\mathrm{acc}(C,0)$ and
$\prrocc(f) = \mathrm{acc}(F,f)/\mathrm{acc}(C,k_{\max})$.
$\prrm$ is the \emph{moving-ceiling} ratio, dividing by the same pipeline at the same depth
with the same re-grounded stage set, and every dose claim rests on it. $\prrd$ divides by the
strict pipeline a practitioner would ship, and is operational only; where it exceeds 1.0, as it
does on two backbones, that is a denominator effect and not the fault restoring clean accuracy
(Table~\ref{tab:headline}, Appendix~\ref{app:deployed}).
$\prrocc$ is the \emph{occlusion} ratio, with one fixed denominator at every visible
fraction $f$. The two matched denominators never share a name, and
Table~\ref{tab:denominators} prints each one as a ratio and as an accuracy.

\paragraph{Blanks remove content, not words.}
The control replaces a fraction of the problem's words with blanks, hash-seeded and nested, so
the visible set at one level is a prefix of the next. Blanks preserve word count, not tokenizer
count, so the ends differ slightly in length: the fully masked arm carries 0.964, 0.904, 1.032
and 0.942 times the prompt tokens of the fully visible one (Table~\ref{tab:denominators}). On
Phi-4 the masked arm is the \emph{longer} one and still scores worse.

\paragraph{Backbones and items.}
We serve four open-weight instruction-tuned backbones locally with thinking disabled
\citep{kwon2023vllm}: Qwen3-14B (primary) and Qwen3-8B \citep{yang2025qwen3}, Phi-4
\citep{abdin2024phi4}, and Llama-3.1-8B-Instruct \citep{grattafiori2024llama3}. Each arm uses 120
\gsm{} items \citep{gao2023pal,cobbe2021gsm8k} at temperature zero (the first 120 in file
order, a fixed non-random subset shared across arms and backbones), plus 20 seeds $\times$ five
arms $\times$ 30 items at temperature 0.7 for the seed-level row. We dropped a fifth candidate
backbone before the pin on a structural trigger (Appendix~\ref{app:registration}). Of the
three registered benchmarks, \gsm{} is the pipeline task and MATH500
\citep{hendrycks2021math} carries the exploratory topology bank; the MMLU bank
\citep{hendrycks2021mmlu} has landed at all 30 topologies with no result computed on it.

\paragraph{A re-run replica sets the jitter yardstick.}
Most headline arms replay from a byte-identical prompt cache (Appendix~\ref{app:replay}); a
replica of the primary backbone's dose arms re-ran under a distinct provider tag no cache could
serve, and disagrees with the original by at most 4 items in 120 on any arm. Every margin below
is compared against that measured jitter, not an imported constant.

\section{The plan protects the procedure, not the thresholds}
\label{sec:plan}

We fixed the analysis plan before computing any inferential statistic, and after the point
estimates of all four backbones' dose curves were already on disk. We do not call it
pre-registered: we protect the inferential procedure, not the thresholds. The analyser held no
paired test, no correction, no bootstrap and no trend test when we pinned the plan; the kill
thresholds, the sidedness, the family membership and the backbone set stay unprotected, and
Appendix~\ref{app:registration} lists them as such beside all eight dated deviations.

Two families are confirmatory: Family~A, the four endpoint contrasts $k = 0$ against $k = 3$, and
Family~B, the four occlusion endpoints. Both use the exact paired test of
\citet{mcnemar1947note}, one-sided, with the correction of \citet{holm1979simple}
within each family at $\alpha = 0.05$ over the registered family size of four, and nothing
corrects \emph{across} the families. We pinned both with all four dose curves already on disk, so
they certify a direction nobody disputed. What they do certify is the procedure: we fixed
interval, correction and effect size first, under a rule these data could have failed.
Appendix~\ref{app:families} carries the rest of the apparatus: the uncorrected rate across the two
families, the interval rules \citep{clopper1934use,wilson1927probable,efron1979bootstrap}, and the
one call the two families share by construction.

We wrote a four-clause kill rule before the result and none of its clauses fired; each would
have killed the pipeline on at least two backbones (Table~\ref{tab:families}). A rule that
cannot fire carries no information, so we simulated it, and
Appendix~\ref{app:families} reports what it would have taken to fire it
(Table~\ref{tab:kill-rule-behaviour}).

\input{tables/T1-headline}

\section{Neither decomposition reliably beats one call}
\label{sec:decompositions}

\paragraph{Is the pipeline worth building? No.}
One call wins on three of four backbones, measured at the \emph{strict} pipeline a practitioner
would deploy. Strict no-fault accuracy at $k = 0$ is 0.567, 0.692, 0.833 and 0.200
against a one-call direct solve at 0.833, 0.817, 0.775 and 0.517: $-0.267$ $[-0.367, -0.167]$,
$-0.125$ $[-0.208, -0.050]$, $+0.058$ $[+0.000, +0.125]$ and $-0.317$ $[-0.417, -0.217]$
(Table~\ref{tab:strict-direct}). The fully re-grounded no-fault pipeline is a kinder comparison:
it scores 0.867, 0.808, 0.825 and 0.383, and separates from the direct solve only on Llama, where
the \emph{re-grounded} arm sits $-0.133$ $[-0.217, -0.050]$ below it
(Table~\ref{tab:mechanism}). Llama's high matched retention is a statement about that ceiling of
0.383, not about robustness.

\paragraph{Pipeline~B ties one call.}
Pipeline~B carries its own matched ceilings at every dose, and every number from it is
exploratory (Appendix~\ref{app:pipelines}).

With no fault, pipeline~B reads $+0.017$ $[-0.017, +0.050]$, $+0.025$ $[-0.025, +0.075]$,
$+0.033$ $[-0.017, +0.083]$ and $-0.050$ $[-0.133, +0.033]$ against the same one-call
reference, item-paired at $n = 120$ (Table~\ref{tab:pipelines}). Every interval covers zero and
the estimates do not agree on a sign, so this is parity, not a win. On the primary the
$-0.267$ deficit becomes a contrast of four discordant items, which erases the loss and leaves
the pipeline level with one call.

\paragraph{The registered pipeline starts lowest where $D$ is smallest.}
With no fault the registered pipeline is the \emph{worst} of the four configurations measured here
on two backbones and third of four on a third, against pipeline~B, full history and a single
call (Tables~\ref{tab:pipelines} and~\ref{tab:history}). Re-grounding repairs a handicap as
readily as it repairs a fault, so that standing feeds straight into $D$: the two smallest
interactions we report are the two backbones where the pipeline starts lowest, and the largest is
Phi-4, where it starts highest.
Appendix~\ref{app:pipelines} works that through backbone by backbone.

\section{The dose-response holds on four of four}
\label{sec:dose}

\begin{figure}[t]
\centering
\includegraphics[width=1.0\textwidth]{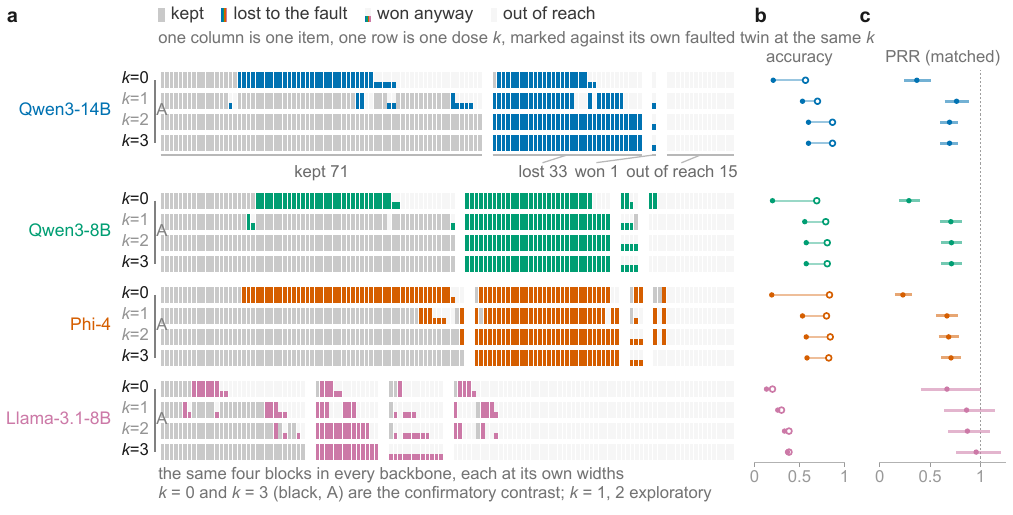}
\caption{Every item in the dose experiment. \textbf{(a)} One column is one problem, one row is
one dose $k$, and a row is the no-fault pipeline at that depth marked against its own faulted
twin: pale means that row answers the item, saturated means the fault takes it away, a notch
means the faulted arm wins it anyway, empty means neither reaches it. The saturated block is the
deficit the dose has to close, and it narrows as $k$ rises on all four backbones; Llama-3.1-8B
has no block to narrow. \textbf{(b)} The two accuracies, filled for the faulted arm, open for its
ceiling. \textbf{(c)} $\prrm$ with 95\% paired item bootstraps. $k = 0$ against $k = 3$ is the
registered confirmatory contrast; $k = 1$ and $k = 2$ are exploratory.}
\label{fig:dose}
\end{figure}

The registered effect size is the
paired difference $\mathrm{acc}(F,k_{\max}) - \mathrm{acc}(F,k_{\min})$ with its exact conditional
interval: $+0.392$ $[+0.303, +0.431]$ on Qwen3-14B and on Phi-4, $+0.375$ $[+0.316, +0.375]$ on
Qwen3-8B, $+0.233$ $[+0.139, +0.289]$ on Llama-3.1-8B (Table~\ref{tab:headline}). Family A holds
on four of four, with Holm-adjusted $p$ from $1.1{\times}10^{-13}$ to $2.1{\times}10^{-6}$.
Appendix~\ref{app:supplementary} explains two printing artefacts in that row.

Grading the dose front-loads the rise. Matched retention on Qwen3-14B runs 0.368, 0.762, 0.692,
0.692, and the other three curves share that shape (Table~\ref{tab:headline};
Figure~\ref{fig:dose}). Qwen3-8B's
$k = 2$ and $k = 3$ arms have identical
correctness vectors and different raw completions, so that arm has saturated.

We claim monotonicity for accuracy under fault, not for the matched ratio. On Qwen3-14B the
matched ratio dips between $k = 1$ and $k = 2$ while accuracy under fault does not fall, because
the ceiling rises from 0.700 to 0.867; Figure~\ref{fig:dose} plots accuracy under
fault against that moving ceiling in its bottom row. Across all twelve adjacent contrasts no adjacent decrease
counts under the registered rule, the largest counted drop being $+0.0$ items against a tolerance
of five.
Decoding noise does not explain the trend either, and Appendix~\ref{app:seeds} reports the
20-seed endpoint test and the rank trend test~\citep{page1963ordered} that say so --- both tests
of \emph{trend} rather than of monotonicity, both over 30 fixed items at non-zero temperature,
and neither a headline interval.

Llama-3.1-8B is a boundary case declared in advance, and it separates the paper's two mechanisms.
Its no-fault accuracy at $k = 0$ is 0.200, and every matched retention interval it produces covers
1, from $[0.407, 1.000]$ at $k = 0$ to $[0.760, 1.204]$ at $k = 3$ --- \emph{including at
$k = 0$}. The fault imposes no detectable cost there at any dose. Yet it is also the backbone
where the pipeline loses most heavily to a single call, by $-0.317$. So on Llama the loss comes
from decomposition itself, and re-derivability governs only the propagation that Llama never
shows. Llama's rejection in Family A therefore speaks to accuracy under re-grounding alone: any
claim that needs a fault to have cost something holds only on the three backbones where one
demonstrably did.

\section{The price is re-derivability, not tokens}
\label{sec:kill}

\begin{figure}[t]
\centering
\includegraphics[width=0.975\textwidth]{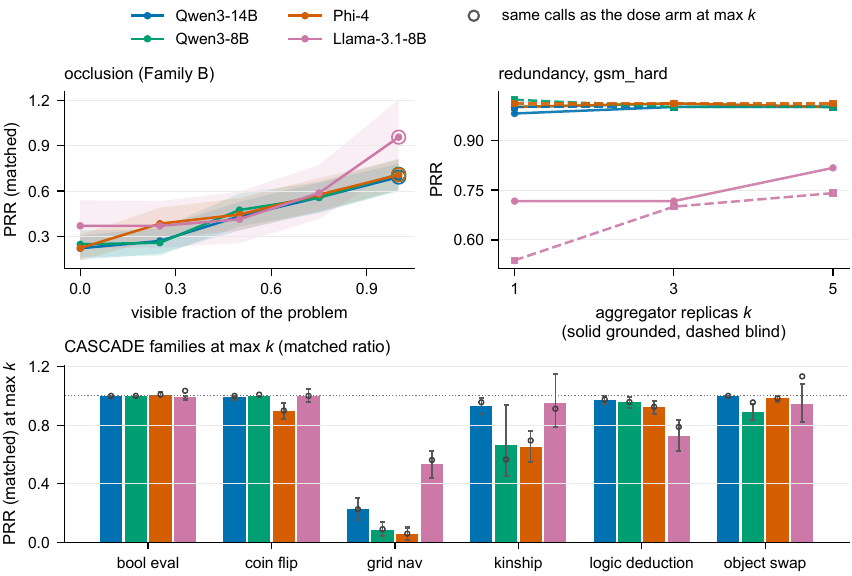}
\caption{The three controls, and what each one rules out. \textbf{Left}: the occlusion
slider, the registered dominant-risk kill test (Family B). The circled point at full visibility is
the same pipeline call as the dose arm at $k_{\max}$, so the two families are not independent
evidence there. \textbf{Right}: redundancy, grounded against blind aggregators at one, three and
five replicas. \textbf{Bottom}: the six cross-family tasks at $k_{\max}$, bars the moving-ceiling
ratio on all 24 cells and hollow markers the deployed ratio beside them. The fault is weaker on
the non-numeric families and one fault across families is not claimed (Section~\ref{sec:flat}).
Bands and bars are 95\% paired item bootstraps.}
\label{fig:controls}
\end{figure}

The registered control kills the token-count explanation: it holds token presence fixed and
varies only derivable content. Across visible fractions 0, 0.25, 0.5, 0.75 and 1.0 the occlusion
ratio $\prrocc$ runs 0.221 to 0.692 on the primary backbone, with paired item bootstrap
intervals of $[0.144, 0.304]$ and $[0.598, 0.781]$ at the two ends. The
other three run 0.247 to 0.711, 0.222 to 0.707 and 0.370 to 0.957, each with its own interval in
Tables~\ref{tab:denominators} and~\ref{tab:families}
(Figure~\ref{fig:controls}, left panel). The curve is
non-decreasing everywhere and strictly rising at every step but one, Llama's first two levels being
an exact tie at 0.370 rather than a rise. Family B holds on four of four,
with Holm-adjusted $p$ from $6.4{\times}10^{-13}$ to $3.7{\times}10^{-6}$; against a registered
kill margin of 0.10 the measured endpoint gaps run from 0.464 to 0.587.

No clause of the kill rule fires, and the breakdown points say how far each was from firing: the
second-smallest deployed gain clears its registered margin by $+28.4$ items and the
second-smallest matched gain by $+13.0$, both far outside the 4-item run-to-run jitter this
campaign measured on the primary backbone and assumed for the other three.
Table~\ref{tab:kill-rule-behaviour} carries every margin against its registered threshold and the
adjacent-decrease clause's own false-kill rate, which is above the nominal level.
Appendix~\ref{app:anchors} adds two exploratory anchors, neither carrying a test.

\section{Re-grounding works by returning the problem}
\label{sec:mechanism}

\paragraph{Does the re-grounded answer use the upstream message?}
The backbones split. The registered
rigging control replaces that message with a fixed uninformative line at $k_{\max}$
(Figure~\ref{fig:mechanism}; Table~\ref{tab:mechanism}). On both Qwen backbones the contrast
covers zero, $-0.025$ $[-0.142, +0.092]$ and $+0.058$ $[-0.050, +0.167]$, so there we
\emph{did not detect} a dependence at $n = 120$. The intervals still admit $\pm 0.117$ and
$\pm 0.108$, and we registered no equivalence margin, so we claim no equivalence. On Phi-4 the
rigged arm scores $+0.208$ $[+0.117, +0.300]$ over the faulty one and on Llama $+0.125$
$[+0.042, +0.217]$: there the \emph{uninformative} upstream beats the faulty one.

\paragraph{Fault-specificity depends on configuration.}
Let $D(k)$ be the re-grounding gain under fault minus the gain without it. On the registered
four-stage pipeline with predecessor-only visibility, $D(k_{\max})$ is $+0.092$ $[-0.025, +0.208]$,
$+0.258$ $[+0.158, +0.358]$, $+0.400$ $[+0.300, +0.508]$ and $+0.050$ $[-0.058, +0.158]$. It
covers zero on Qwen3-14B and on Llama-3.1-8B. \emph{In that configuration} re-grounding helps
the no-fault pipeline about as much as the faulty one, and the interior doses agree
(Table~\ref{tab:mechanism}). A null here is not an absence: this design detects an interaction
of 0.180 on Qwen3-14B and 0.160 on Llama at 80\% power. The qualifier matters, because the
same estimand excludes zero on four of four under pipeline~B and on the primary under full
history (Section~\ref{sec:pipelines}), so every statement here that $D$ covers zero on the
primary is about pipeline~A with predecessor-only visibility and nowhere broader.

\paragraph{Grounding the inspector wins four of four.}
\citet{huang2025resilience} report an Inspector agent recovering most of the loss
without varying what it can see. We ran that contrast at the headline standard
(Table~\ref{tab:defense}). Grounding the inspector is worth
$+0.608$ $[+0.517, +0.700]$, $+0.575$ $[+0.483, +0.667]$, $+0.608$ $[+0.500, +0.708]$ and $+0.358$
$[+0.267, +0.458]$ over a blind one, and $+0.608$,
$+0.575$, $+0.600$ and $+0.283$ over the strict pipeline. Every interval excludes zero. On
the two Qwen backbones those are one measurement rather than two: the blind inspector's per-item
vector is identical to the strict pipeline's, changing no item at all. Run
on a \emph{healthy} pipeline with no fault to find, it moves accuracy against
that healthy pipeline by $+0.000$, $+0.025$
$[+0.000, +0.058]$, $-0.192$ $[-0.267, -0.117]$ and $-0.133$ $[-0.200, -0.075]$. The one nominal
gain is not separable from zero at $p = 0.250$. The two intervals that exclude zero are both
costs to a healthy pipeline, not deficits against one call. An ungrounded inspector is not a free
safety net but a stage that can break a working pipeline.

Which ceiling the ratio divides by decides how close it comes: against the isomorphic no-fault
cell the grounded arm reaches 0.980 $[0.931, 1.030]$ on the primary, and against a cell at a
\emph{different} stage count the same arms run as
high as 6.250 $[3.500, 16.667]$ --- a defence apparently scoring six times its ceiling is
describing its denominator (Appendix~\ref{app:defense}).

\section{The effect survives both other configurations}
\label{sec:pipelines}

\paragraph{Pipeline~B's interaction excludes zero.}
On pipeline~B (Section~\ref{sec:decompositions}), matched retention climbs from 0.137 to 0.647 on Qwen3-14B, 0.040 to 0.293 on
Qwen3-8B, 0.196 to 0.531 on Phi-4 and 0.232 to 0.593 on Llama (Figure~\ref{fig:pipelines}b;
Table~\ref{tab:pipelines}). The interaction is $+0.433$ $[+0.342, +0.525]$, $+0.225$ $[+0.142,
+0.308]$, $+0.275$ $[+0.192, +0.358]$ and $+0.158$ $[+0.042, +0.275]$: it excludes zero on four
of four, against two of four on pipeline~A. Pipeline~B's no-fault gain on the primary is exactly
$0.000$, because re-grounding a pipeline with nothing to recover from changes
nothing at all, so the whole of its $+0.433$ is fault-specific. Its no-fault ceiling is also higher
than pipeline~A's (0.850 against 0.567): there the three-stage decomposition is both the better
pipeline to deploy and the one where re-grounding is most clearly a fault remedy.

\paragraph{Full history changes nothing under fault, or hurts.}
\label{sec:history}
We ran the same pipeline with full history on all four backbones (Table~\ref{tab:history}): the
lever survives where information is plentiful, and the scarcity objection fails. At $k_{\max}$
the full-history arm differs from the predecessor-only arm by $+0.000$ $[-0.042, +0.042]$,
$+0.008$ $[-0.033, +0.050]$ and $-0.033$ $[-0.075, +0.000]$ on the first three backbones, none
distinguishable. On Llama it is \emph{worse}, by $-0.075$ $[-0.133, -0.017]$, $p = 0.022$, which
is 9 of 120 items against a within-campaign jitter of 4. The dose response survives on all four
(Table~\ref{tab:history}), with the interaction reading $+0.317$ $[+0.200, +0.433]$, $+0.258$
$[+0.150, +0.359]$, $+0.367$
$[+0.267, +0.467]$ and $+0.033$ $[-0.083, +0.150]$. So the strict setting is not the source of the
effect, and on the primary the interaction that covers zero under predecessor-only visibility
excludes zero here, on an exploratory two-sided test. This arm passes every earlier stage's
\emph{output}, never the initiating problem statement. AutoGen's group chat and ChatDev's phase
templates carry the task as well, and a MetaGPT role sees the pool filtered by its watch set, so a
real framework default sits between this arm at $k = 0$ and at $k_{\max}$
(Table~\ref{tab:history}); we report both bounds and claim neither is the default.

\paragraph{The first stage buys almost the whole gain.}
At the headline standard the first re-grounded stage returns $+0.394$ of matched retention on
$+59.8$ tokens per item against a 917.5-token pipeline, and the two after it return $-0.070$ and
$+0.000$ for a further 156 tokens, the fall being the rising ceiling again
(Table~\ref{tab:cost-headline}, Figure~\ref{fig:mech}D3). The shape holds on the other three
backbones. Tokens rise monotonically with $k$ on every backbone.

\section{What the controls decide, and where the claims stop}
\label{sec:flat}

\paragraph{The collapse is family-specific.}
Six non-mathematical reasoning families, 120 items per family per backbone, run through the same
injector (Figure~\ref{fig:controls}, bottom panel). State-tracking families collapse and then
recover, while binary-answer families barely move, because a two-valued answer space bounds how
far one wrong intermediate can push them.
The other families, all 48 matched ceilings and the two ratios are in
Appendix~\ref{app:cascade} (Table~\ref{tab:cascade}).

\paragraph{Two verdicts turn on the control.}
A one-line replacement is shorter as well as emptier, so it confounds ``nothing to derive from''
with ``less to read'', the distraction account the literature would predict
\citep{shi2023distracted}. A post-pin exploratory arm replaces the upstream message with a
digit-free filler of exactly its character length, and it decides two of the four verdicts. On the
primary backbone it refutes the distraction account outright: the equal-length filler scores 0.408,
\emph{below} both the one-line control ($-0.167$ $[-0.250, -0.083]$) and the faulty message itself
($-0.192$ $[-0.308, -0.067]$), so content removed at fixed length does cost accuracy and the
one-line null was an artefact of brevity.

The control decides the verdict, so we name it with every one. Under the \emph{post-pin
length-matched} control we detect a dependence on the primary, none on Qwen3-8B at $+0.025$
$[-0.083, +0.133]$, and one in the anomaly direction on Phi-4 at $+0.242$ $[+0.150, +0.333]$; on
Llama the one-line anomaly disappears once length matches, at $+0.058$ $[-0.033, +0.150]$
(Table~\ref{tab:mechanism}). We rest on the better-controlled one.

\paragraph{Where every claim here stops.}
Every task is verifiable reasoning with a checkable answer; every item-level inference holds only
on the 120 fixed \gsm{} items these arms share; every backbone is open-weight, instruction-tuned and
served with thinking disabled; one fault at one injection site is not a severity sweep; and
everything outside the two registered families is exploratory.

\paragraph{Redundancy and topology search both fail.}
Appendix~\ref{app:nonlevers} reports both exploratory nulls in full.

\section{Related work}

\paragraph{Fault injection and failure taxonomies.}
Large trace-corpus taxonomies of multi-agent failure are observational: they name what breaks
without manipulating a candidate cause \citep{cemri2025why,albayaydh2026beyond}. The closest
instrumented work injects a catalogue of fault types into several systems
\citep{jia2026masfire} and reports that closed-loop designs neutralise a large share of the faults
that collapse linear workflows, explaining the effect by a shared message pool downstream agents
retrieve from. That explanation is stated descriptively and never tested; this paper manipulates
it as a hypothesis. That work is \emph{prior}, not concurrent, on
many fault types against our one. Adjacent efforts instrument observability
for agentic pipelines
\citep{seyedghorban2026observability}, transfer error recognition across agents
\citep{yu2026correct}, and bound reliability analytically \citep{ao2026reliability}.

\paragraph{The closest prior measurement.}
\citet{huang2025resilience} share our fault-injection setting. Their Inspector prompt
passes the chat history, so it is \emph{grounded} by construction, and because they never vary
what it can see they cannot separate adding an inspector from giving one the problem
(Section~\ref{sec:mechanism}).

\paragraph{Re-derivability as a named variable elsewhere.}
\citet{kwon2026reclaim} named the variable first, as the quantity deciding whether a lossy
memory helps or hurts a \emph{single} agent. Ours is the manipulation inside a multi-stage
pipeline, with a matched ceiling at every dose and an occlusion control, where the object is a
message between stages. Neighbouring work measures what is lost in an isolated context
\citep{radhakrishnan2023decomposition}, how position in a long context degrades use of it
\citep{liu2024lostmiddle}, how far a decomposed prompt carries a multi-step problem
\citep{zhou2023leasttomost,dziri2023faithfate}, and how skill libraries are refactored for
reconstruction \citep{li2026skillpseudocode}. Two results bound what our contrasts can mean: a
confidently stated irrelevant quantity costs accuracy in proportion to its salience
\citep{shi2023distracted}, and a stated chain of reasoning need not be the computation that
produced the answer \citep{lanham2023faithfulness}.

\paragraph{No widely used framework defaults to strict.}
Strict predecessor-only visibility maximises our lever and no widely used framework defaults to
it: AutoGen carries the message history \citep{wu2024autogen}, MetaGPT routes through a
publish-subscribe pool \citep{hong2024metagpt}, ChatDev hands forward more than the preceding
artefact \citep{qian2024chatdev}. We measure the strict setting because it isolates the
variable, and Section~\ref{sec:pipelines} measures the full-history one too. A large lane
instead optimises the communication graph
\citep{zhuge2024gptswarm,zhang2025gdesigner,zhang2025maas,zhang2025aflow,wu2026card,zeng2026sigma,jiang2026topologydiffusion,tastan2026selforg,wang2026rumad,wu2026dmoa,wang2025agentdropout,leong2025amas},
with protocol-layer remedies alongside \citep{mao2025protocol};
Section~\ref{sec:flat} reports what optimising the graph bought on the one bank we measured.

\paragraph{Byzantine and adversarial framings.}
Our setting inverts the threat model of \citet{lamport1982byzantine}, where the
faulty component is adversarial and the remedy is agreement, a framing recent work carries into
language-model collectives
\citep{lee2026byzantine,elmir2026cheaptalk,jo2025decentralized,mao2025ibgp,wang2026btrag} and that
\citet{bappy2026adversarial} extend to architecture. Self-correction
\citep{huang2024selfcorrect,tyen2024errorlocation,madaan2023selfrefine} and
redundancy-as-reliability \citep{wang2023selfconsistency,du2024debate} bound what a verifier can do
without external grounding, the boundary our blind-inspector row runs into, and error-propagation
models \citep{singh2026snowballpropagation,zhang2024snowball,konstantinou2026coding,nguyen2025socialcost}
describe the cascade dynamics we intervene on.

\section{Conclusion: ground the stage that lost the information}
\label{sec:discuss}

\paragraph{What the evidence supports.} One manipulated variable moves fault retention on four of
four backbones, with a matched ceiling at every dose and a registered control separating
derivable content from token presence. Full message history reproduces it on four of four, so
the setting we registered does not manufacture the effect
(Section~\ref{sec:history}).

\paragraph{What it does not.} We missed the registered transfer target of 0.900: among the three
eligible backbones the third-largest matched retention is 0.692, a disclosed deviation. The
interaction covers zero on two of four backbones under the registered predecessor-only pipeline,
and that it excludes zero under the other two configurations is evidence about the configuration
rather than a repair (Section~\ref{sec:mechanism}). On Llama the fault has no detectable cost at
any dose, so that backbone cannot test a claim about what a fault costs.
Section~\ref{sec:flat} states where every claim stops.

\paragraph{Full message history is not re-grounding.}
When the fault lands at the first stage the history \emph{is} the contamination, so passing more
of it passes more of the fault. Re-grounding on top of full history still lifts accuracy under
fault on four of four (Section~\ref{sec:history}), so the framework default is the setting where
this remedy still applies and still works.

\paragraph{For practitioners.} The pipeline loses to a single direct call on three of four
backbones and never reliably wins. Passing the full history narrows that deficit on one of the
three without closing it, and the narrowing is a change in the pipeline's own accuracy rather than
in the remaining gap to one call (Table~\ref{tab:strict-direct}). Teams that run a pipeline anyway,
for the tools, budgets or policies one call cannot carry, should ground the stage that lost the
information, and ground it first (Section~\ref{sec:pipelines}). The prediction this
leaves open is falsifiable: partial re-derivability should yield partial retention, the shape the
occlusion interior measures, and it extends to open-ended generation, multi-hop retrieval and
tasks with no checkable answer, all untested here.

\clearpage
\label{lastmainpage}%
\label{firstpageafterbody}

\subsection*{AI use statement}

Generative AI tools were used solely to assist with language editing and improve the clarity and
readability of the manuscript. All scientific content, analyses, results, interpretations, and
conclusions were developed and verified by the authors, who take full responsibility for the
content of the paper.

\subsection*{Ethics statement}

This paper measures how a staged pipeline of language-model agents recovers from an injected
fault, on public reasoning benchmarks with open-weight models served locally. It introduces no new
model, dataset or human-subject data, and the only data it shares, available from the corresponding
author on request, are per-item correctness vectors, token counts and model outputs for public
benchmark problems. The foreseeable effect is a modest reliability
improvement at a small token cost. Two risks matter. A measured robustness gain invites
over-trust, and our own comparison
against a single call shows the architecture we harden is worse than not building it on three of
four backbones. And our fault is deterministic, single-site and benign by construction, so nothing
here speaks to an adversary who chooses it.

\subsection*{Reproducibility statement}

Appendix~\ref{app:repro} is the reproduction entry point. It names the result files each headline
number is read from, the analyser that derives every point estimate and interval from them, the
gate suite that re-checks them, and the single command that rebuilds every table and figure in
this paper from those files. Appendix~\ref{app:registration} states when the analysis plan was
fixed relative to the data, and lists all eight dated deviations from it, so a reader can tell
which quantities were protected by the plan and which were not; Appendix~\ref{app:replay} records,
arm by arm, whether an arm was replayed from a byte-identical prompt cache or re-measured, and
reports the run-to-run jitter of a replica lane that could not be served from cache.
Section~\ref{sec:setup} fixes the items (the first 120 \gsm{} items in file order, one fixed
subset shared by every arm and backbone), the four open-weight backbones, and the decoding
settings; Appendix~\ref{app:denominators} prints both matched denominators as ratios and as
accuracies so no reported ratio has to be reconstructed from a mean. Every per-item correctness
vector behind an item-paired test is in the result files, available from the corresponding author
on request, which is what makes the paired
intervals recomputable rather than merely re-readable.

\bibliographystyle{iclr2027_conference}
\bibliography{references}

\newpage
\appendix
\input{appendix}

\end{document}

%% file: authors.tex
\author{Tianqi Bu$^{1,2}$\thanks{Corresponding author: \texttt{tianqi.bu@rutgers.edu}} \quad
YuXuan Peng$^{3}$ \quad Junteng Tu$^{4}$ \quad Henghui Xiao$^{1}$ \\
$^{1}$Rutgers University \quad $^{2}$Nanjing Tech University \\
$^{3}$Nanjing University of Posts and Telecommunications \quad $^{4}$New York University}

%% file: tables/T1-headline.tex
\begin{table}[t]
\centering
\footnotesize
\caption{Per-backbone headline. Clean ceilings are the matched no-fault cells; PRR (matched) divides by the ceiling at the same $k$, and PRR (deployed) by the deployed strict pipeline at $k{=}0$, which is an operational number only. $\Delta$acc is the registered paired difference with its exact conditional interval; $p$ is Holm-adjusted within Family A.}
\label{tab:headline}
\begin{tabular}{lrrrlllrl}
\toprule
backbone & $n$ & clean & clean & PRR (matched) & PRR (deployed) & $\Delta$acc & Holm $p$ & retention \\
 &  & min $k$ & max $k$ & min to max $k$ & min to max $k$ & [95\%] & family A & vs.\ ceiling \\
\midrule
Qwen3-14B & 120 & 0.567 & 0.867 & 0.368\,$\to$\,0.692 & 0.368\,$\to$\,1.059\,($>$1) & +0.392 [+0.303, +0.431] & 8.3e-12 & below \\
Qwen3-8B & 120 & 0.692 & 0.808 & 0.289\,$\to$\,0.711 & 0.289\,$\to$\,0.831 & +0.375 [+0.316, +0.375] & 1.1e-13 & below \\
Phi-4 & 120 & 0.833 & 0.825 & 0.230\,$\to$\,0.707 & 0.230\,$\to$\,0.700 & +0.392 [+0.303, +0.431] & 8.3e-12 & below \\
Llama-3.1-8B & 120 & 0.200 & 0.383 & 0.667\,$\to$\,0.957 & 0.667\,$\to$\,1.833\,($>$1) & +0.233 [+0.139, +0.289] & 2.1e-6 & indist. \\
\bottomrule
\end{tabular}
\vspace{0.4em}
\begin{minipage}{\linewidth}\footnotesize
Retention labels are the registered wording. \emph{below}: the paired-bootstrap interval lies below the matched ceiling. \emph{indist.}: it covers 1, and the row is reported as retention indistinguishable from the matched ceiling, never as absorption and never as an anomaly. ($>$1) marks a PRR (deployed) point estimate above 1: a denominator effect to explain, never a claim that the faulty arm beats clean.
\end{minipage}
\end{table}

%% file: appendix.tex
\section{The registration, and every deviation from it}
\label{app:registration}

\paragraph{When the plan was fixed, stated exactly.}
The analysis plan was fixed before any inferential statistic was computed, and \emph{after} the
point estimates of all four backbones' dose curves were already on disk. The record is: all four
dose curves landed and were tabulated on 2026-09-12, and the registration was written later the
same day. The one-backbone clause attaches only to the kill \emph{thresholds}, which descend from
a design document written on 2026-09-11, when one backbone's curve was known. We do not call any
of this pre-registered.

What the plan protects is the inferential procedure: the analyser held no paired test, no
multiplicity correction, no bootstrap and no trend test at the moment the plan was pinned, so no
interval, test or correction in this paper was computed on landed data before the procedure that
produces it was written down. What it does \emph{not} protect, so that a reader can discount it:
the kill thresholds, the sidedness of each test, the membership of the two confirmatory families,
the backbone set, and which contrast is confirmatory. The endpoint outcome of Family A was not in
doubt when the plan was pinned; the interval, the correction and the exact effect-size procedure
were.

\paragraph{Deviations and pre-pin history: all eight, dated.}
Items 1--7 are recorded in the project's deviations file as pre-pin history: they are what
changed between the design document and the pinned registration. Item 8 was found after the
analysis and is disclosed here.

\begin{enumerate}\itemsep2pt
\item \textbf{Backbone swap} (2026-09-11). A fifth candidate backbone was replaced by
  Llama-3.1-8B-Instruct on a structural trigger and not on an outcome: its chat template
  ignored the thinking-disable parameter and returned 13--21 empty completions per experiment,
  which the result writer refuses at write time rather than scoring an empty string as a wrong
  answer. The replacement weights are a byte-identical ungated mirror of the released model; we
  cite the model \citep{grattafiori2024llama3} and record the mirror as provenance.
\item \textbf{Transfer target rewritten} (2026-09-12). The design document's transfer target
  (retention at $k_{\max}$ at least $0.9 \times$ clean, among backbones with retention below 0.5
  at $k_{\min}$, on at least three of four) measured 1 of 4 on the deployed ratio and was
  unmeetable on the matched ratio. It was never part of the kill rule. ``Restores to clean'' was
  withdrawn, and the claim became ``raises retention by at least 0.20 on both ratios, monotone, on
  $m$ of 4'', with $m$ as measured.
\item \textbf{A kill clause removed} (2026-09-12). The design document's one-sided
  $z \ge 2.5$ clause was removed from the decision rule: it duplicated the McNemar family at a
  different $\alpha$ with an undefined null.
\item \textbf{Monotonicity rule changed} (2026-09-12). An adjacent decrease is judged by an
  item-paired one-sided 95\% lower bound rather than by a 0.03 ratio tolerance, which at $n = 120$
  is 0.7--3 items and therefore below the measured temperature-zero jitter. The 0.03 flag stays,
  labelled descriptive.
\item \textbf{Occlusion endpoint redefined} (2026-09-12). The fully masked level returns every
  word as a blank, which holds token presence while removing content; the pre-amendment code
  returned an empty string, which would have confounded the control with a shorter prompt. Masks
  are hash-seeded and nested, and all four occlusion arms were re-run under the corrected
  definition before the kill clause was evaluated.
\item \textbf{Seed floor scoped} (2026-09-12). The 20-seed cell had been measured on the primary
  backbone only; the campaign was changed to run it in every per-model lane, and until all four
  landed the seed claim was scoped to the backbones measured. All four have since landed.
\item \textbf{Fault type re-run} (2026-09-12). The redundancy and cross-family drivers had run the
  script default (\texttt{message\_truncation}); before any figure pairs them with the dose curve
  they were re-run with \wi{}.
\item \textbf{Eligible-set shortfall} (found 2026-09-15, after the analysis). The registered
  transfer rule takes the third-largest $\prrm(k_{\max})$ among the backbones whose
  $\prrm(k_{\min})$ is below 0.5. Four backbones were registered as the pool; only three are
  eligible, because Llama-3.1-8B's $\prrm(k_{\min})$ is 0.667 and it is filtered out. The
  delivered threshold of 0.692 is therefore the third-largest over a pool of three rather than of
  four, which is a weaker statement than the registration intended. It is recorded as a dated
  post-pin disclosure row in the project's deviations file alongside the seven pre-pin rows above;
  no registered formula or threshold changed.
  The registered target of 0.900 was not met either way.
\end{enumerate}

\paragraph{The last registered cell, and what it does not buy.}
All three registered standard benchmarks for the exploratory flat-regime scripts are now covered:
\gsm{} is the pipeline task every headline arm runs on, MATH500 \citep{hendrycks2021math} carries
the topology bank the rows above are computed from, and the MMLU bank \citep{hendrycks2021mmlu}
has completed all 30 of its topologies (Table~\ref{tab:flat}). That coverage is all we report from
it. No statistic here is computed on the MMLU bank, so the flat-regime result stays what it was:
one bank, on one backbone, exploratory throughout.

\section{Per-arm replay status, and the measured run-to-run jitter}
\label{app:replay}

Every driver persists, for every arm, the per-item correctness vector, the parsed prediction, the
final-stage raw output, the prompt and completion token counts, and the per-arm server-call and
cache-hit counters. An arm counts as a \emph{replay} only when its per-arm server-call counter is
zero, meaning its per-item vector was recomputed from a byte-identical prompt cache rather than
re-sampled; otherwise it is a fresh temperature-zero measurement.

The status is now known for every backbone. Qwen3-14B: 8 of 8 dose arms replay at zero server
calls, as does its full twenty-seed cell. Qwen3-8B: 8 of 8 dose arms replay at zero server
calls. Phi-4 and Llama-3.1-8B: 6 arms replay and 2 are fresh, and the fresh arms are
their re-grounded no-fault cells.

A re-score is recorded separately from a replay, and it happened on the primary backbone rather
than a secondary one. When an arm stored only a mean, the analyser recomputes it per item from the
byte-identical prompt cache at zero server calls and keeps the old value; that is a re-score, not a
re-measurement. Eight of Qwen3-8B's arms were recomputed this way and \emph{none} of them moved:
every delta is exactly $0.000$. Four of Qwen3-14B's moved, each by the same $-0.017$, which is two
items of 120: \texttt{clean} from $0.583$ to $0.567$, and \texttt{reground\_k1},
\texttt{reground\_k2} and \texttt{reground\_k3} alongside it. We name the backbone and the arms
because two of the four, \texttt{clean} and \texttt{reground\_k3}, are the endpoints of the
contrast this paper's headline rests on, and every value printed here is the re-scored one. A
two-item move is half the measured run-to-run jitter of 4 items and changes no verdict; being
recomputed and having moved are different properties, and only the second one is this.

\paragraph{The shared endpoint is arithmetic, not agreement.} The occlusion arm at full visibility
and the dose arm at $k_{\max}$ are the same calls under a shared provider tag, so their per-item
vectors agree on 120 of 120 items \emph{by construction}. That agreement is never reported as a
reproducibility result, and it is the reason the two confirmatory families are not counted as
independent evidence about that arm.

\paragraph{The jitter, now measured rather than imported.} One arm set was re-run precisely so the
breakdown distances would have a within-campaign yardstick. A replica of the primary backbone's
eight dose arms ran fresh at temperature zero under a distinct provider tag, whose cache keys
differ so the job could not be served from cache. Against the original arms it disagrees on 4 of
120 items at \texttt{clean}, 2 at \texttt{clean\_k1}, 1 at \texttt{clean\_k2}, 0 at
\texttt{clean\_k3}, and 1, 3, 1 and 1 across \texttt{reground\_k0} to \texttt{reground\_k3}: 13
item flips across 8 arms, with a largest per-arm disagreement of 4. The corresponding accuracy
deltas are at most $+0.017$. The analysis plan carried an imported figure of 1 to 2 items from two
earlier jobs; the measured value is larger, so the main text uses 4 and
the breakdown distances of $+28.4$ and $+13.0$ items clear it by a wide margin either way. We
report the larger number because a jitter estimate that flatters the conclusion is worth nothing.
The full per-arm table, carrying the counters and the replay label for every arm on every
backbone, is available with the code from the corresponding author on request.

\input{tables/T2-families}
\input{tables/T2b-kill-rule-behaviour}

\section{Which items each dose recovers}
\label{app:atlas}

\begin{figure}[t]
\centering
\includegraphics[width=1.0\textwidth]{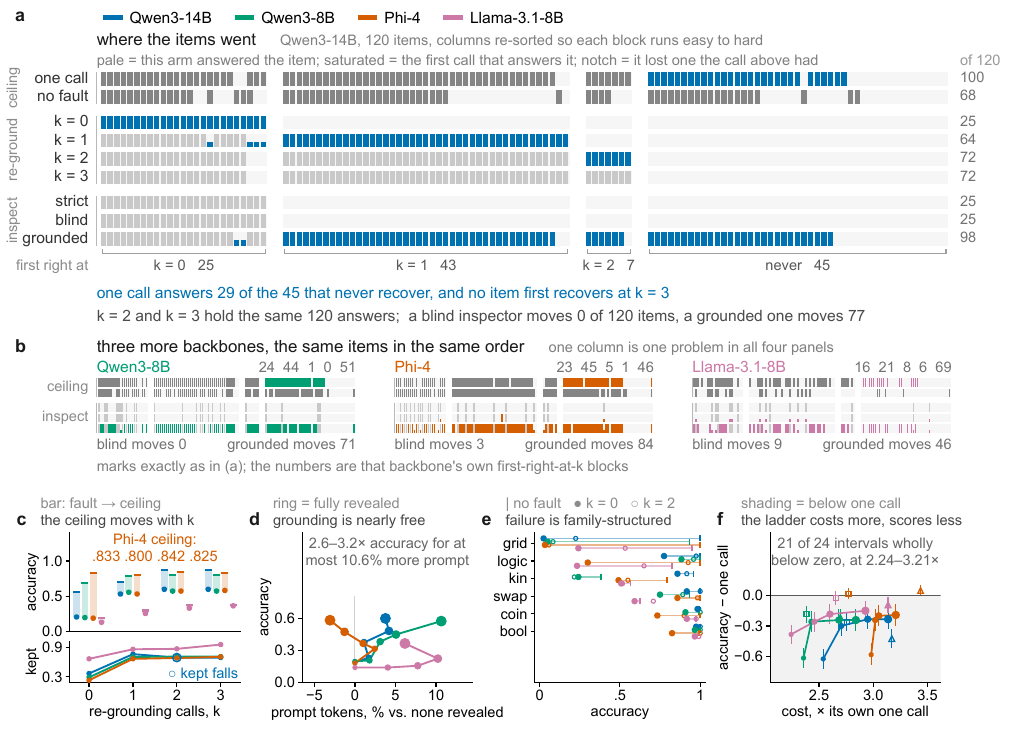}
\caption{Where the items went. Hue is the backbone and nothing else. Inside a raster a grey
cell is one item answered right (darker in the two reference rows), and chroma marks what an
arm did that its own comparison did not: on the dose ladder, the call that first answers an item;
on the inspector rows, what that inspector wins against the strict pipeline; on the single-call
row, the items it answers that no dose of re-grounding does. A tab at the top of a cell is an
item lost to that same comparison. \textbf{(a)} Qwen3-14B's 120 items ordered by the first $k$ at
which its faulted pipeline is right, and none is first right at $k=3$. \textbf{(b)} The reference
and inspector rows only, in the hero's column order, so one column is one problem in all four
rasters; the dose ladder is drawn once, in (a). \textbf{(c)} Faulted accuracy against the matched
no-fault cell at the same depth, and below it the fraction of its own ceiling each arm keeps. The
ceiling moves with $k$, so retention can fall where accuracy rises (the ringed point is the one
arm where it does), and on Phi-4 the ceiling is not even monotone. \textbf{(d)} Accuracy against
the prompt budget it is bought with; on Phi-4 the prompt gets \emph{shorter}, by $-3.1\%$.
\textbf{(e)} The six families: no fault, and the two faulted doses the panel draws. 15 of the 72
family--backbone--arm cells sit at accuracy 1, and 5 of the 48 faulted cells land above their own
no-fault arm. \textbf{(f)} Each arm against its own single call, with paired bootstrap intervals:
circles are the dose ladder, the open square and triangle the two rigging controls. The 21 lying
wholly below zero span $2.2$ to $3.2\times$ the cost; all three exceptions are controls, one of
them at $3.4\times$.}
\label{fig:hero}
\end{figure}

Figure~\ref{fig:hero} and every statement in this section sit outside both confirmed families,
which the registration confines to the four endpoint contrasts and the four occlusion endpoints
(Appendix~\ref{app:registration}): all of it is exploratory and carries no test.Ratios and discordant counts each compare one pair of arms. A matched ratio reports how much of
its own moving ceiling an arm keeps, and a McNemar discordant count reports how many items moved
and in which direction (Section~\ref{sec:dose}). Both leave out which items. The discordant count
between $k = 0$ and $k = 3$ reads the same whether the wins at $k = 1$ and at $k = 2$ land on one
set of problems or on two disjoint ones. It reads the same again whether each arm's correct set
contains the arm below it or matches it only in size. Figure~\ref{fig:hero} draws that identity.

The figure scopes its dose ladder to one backbone. Panel~(a) sorts Qwen3-14B's 120 columns by the
first dose that answers each item, and only panel~(a) draws the four dose rows. Panel~(b) reuses
that column order for the other three backbones, drawing their reference and inspector rows, so
the same column position names the same problem in every raster. Each miniature prints its own
first-right counts above it. Every item-level number below is the primary backbone's.

Sorted that way, the ladder front-loads. 25 of the 120 items are already right at $k = 0$, 43 more
turn over at $k = 1$ and 7 at $k = 2$, and the $k = 3$ block is empty. An empty $k = 3$ block
establishes only that the last dose wins no item the first three missed, and losses stay possible
under it, so it is weaker than the flat endpoint the matched ratio reports. The check that closes
that gap is the one Section~\ref{sec:dose} runs for Qwen3-8B: compare the arms' per-item
correctness vectors instead of their means. Here \texttt{reground\_k2} and \texttt{reground\_k3}
agree on all 120 items, as panel~(a) states under the raster, so the endpoint on this backbone is
flat item by item as well as on average.

The faulted ladder leaves 45 items unanswered at every dose, and one direct call answers 29 of
them. That pair sets a faulted four-stage pipeline against a single unfaulted call, so it reports
what one clean call reaches among the items no dose of the faulted ladder reaches. It is a
different comparison from the architecture result in Section~\ref{sec:decompositions}, which sets
the \emph{no-fault} pipeline against that same call; the two share the reference arm and differ in
the manipulation. Panel~(a) draws the one-call and the no-fault rows beside the ladder and
panel~(b) draws them for the other three backbones, so the no-fault comparison is on the page as
items even though no number here states it.

\section{Families, the kill rule, the breakdown points and the rule's power}
\label{app:families}

Table~\ref{tab:families} carries both confirmatory families and the four kill clauses with the
set of backbones failing each. Table~\ref{tab:kill-rule-behaviour} carries how that rule behaves:
the breakdown margins, the measured run-to-run jitter, the transfer threshold, the adjacent
clause's false-kill rate and the rule's simulated power. The
breakdown margins are the informative form of a sensitivity analysis here: rather than re-running
the rule inside an arbitrary band around each registered threshold, each row reports the smallest
margin at which that clause would have fired, so the reader can place any threshold they prefer
against the measured value.

\paragraph{The rest of the inferential apparatus.} Nothing corrects across the two confirmatory
families, so the nominal rate over all eight tests is about 0.098 rather than 0.05. It changes no
decision here, because every adjusted $p$ falls below $10^{-5}$. Family~A effect sizes are exact
conditional intervals on the discordant pairs \citep{clopper1934use}; per-arm intervals are
Wilson intervals \citep{wilson1927probable}; every ratio interval is a paired item bootstrap over the item
index \citep{efron1979bootstrap} with $B = 10{,}000$ resamples. Occlusion at full visibility with
the fault is the same call as dose $k = 3$, so the two families share that measurement \emph{by
construction}: their zero item disagreement is an identity and bounds nothing.

\paragraph{A rule that did not fire, and its power.} ``Not killed'' carries information only if
the rule could have fired. Simulated over the landed per-item vectors, the deployed-gain clause
fires with probability 0.999 against a pipeline with no dose response at all and 0.705 at the
registered margin itself, which is the behaviour a boundary case should have; the largest true
gain it still detects at 80\% power is 0.150; and at the gains this campaign measured it fires
with probability 0.000. The rule had every chance to fire here and did not.

\paragraph{The transfer row.} This is the row that did not go our way, and it is reported as
measured. The registered quantity is the third-largest $\prrm(k_{\max})$ among the backbones
eligible under the registered filter, which is those whose $\prrm(k_{\min})$ is below 0.500. Three
of the four are eligible; Llama-3.1-8B, at 0.667, is not. The third-largest among the eligible
three is 0.692, against a registered target of 0.900. All four backbones do clear 0.692. The
shortfall in the eligible set is item 8 of Appendix~\ref{app:registration}.

\section{Two descriptive anchors}
\label{app:anchors}

Both are read from the preflight file on the primary backbone, both are exploratory, and neither
carries a test. Leaking the original problem at the injection site lifts accuracy to 0.600,
against 0.567 unfaulted and 0.208 under the fault; that arm is the $k = 3$ arm under another name,
so it is an anchor rather than a control and the analysis plan says so. Damage varies with the
injection site and is \emph{not} ordered by depth: injecting the same fault at the first, second,
third and fourth stage gives 0.208, 0.200, 0.325 and 0.017, so the two extremes are the shallowest
and the deepest site while the third is the mildest of the four.

\section{Seed-level tests}
\label{app:seeds}

Twenty seeds, five arms and 30 items per seed-arm at temperature 0.7 on every backbone, with arms
paired by seed. The five arms are the four doses $k = 0,\dots,3$ under fault plus the no-fault
cell; the trend tests below order the four doses, and the no-fault arm is the ceiling they are
read against, which is why the text elsewhere speaks of four ordered arms. The endpoint paired $t$ (df 19, one-sided) is 22.8 on Qwen3-14B, 50.4 on Qwen3-8B,
27.0 on Phi-4 and 9.9 on Llama-3.1-8B, rejecting in all four cases. The per-seed least-squares
slope, tested as a one-sample one-sided $t$ on the same degrees of freedom, gives 21.8, 46.6, 25.3
and 9.4. Page's rank trend statistic~\citep{page1963ordered} over 20 blocks of four ordered arms
gives 590.5, 564.5, 558.5 and 580.5 against a maximum of 600.

Two labels matter. First, the per-seed slope test and Page's statistic are \emph{trend} tests, not
monotonicity tests: they can pass on a curve with a local dip. The monotonicity claim in the main
text rests only on the item-level adjacent-decrease rule, which is evaluated per adjacent pair and
not on a fitted slope. Second, these intervals quantify decoding variability over 30 fixed items
at non-zero temperature; they are not the headline interval, and the headline intervals are the
item-level ones at temperature zero on 120 items.

\section{Supplementary intervals}
\label{app:supplementary}

The registered Family A interval is the exact conditional one on the discordant pairs
\citep{clopper1934use}. The paired item bootstrap is supplementary and is recorded here so that
the two can be compared: $[0.300, 0.483]$ on Qwen3-14B, $[0.292, 0.458]$ on Qwen3-8B, $[0.292,
0.492]$ on Phi-4 and $[0.142, 0.325]$ on Llama-3.1-8B, against exact intervals of $[0.303,
0.431]$, $[0.316, 0.375]$, $[0.303, 0.431]$ and $[0.139, 0.289]$. The discordant-pair counts that
the exact interval conditions on are $b = 3, c = 50$ on Qwen3-14B and on Phi-4, $b = 0, c = 45$ on
Qwen3-8B and $b = 5, c = 33$ on Llama-3.1-8B. Qwen3-8B's $b = 0$ is why its exact upper limit
equals its point estimate, and the identical Qwen3-14B and Phi-4 rows follow from identical
marginal tables on different per-item vectors.

One further supplement is available with the result files from the corresponding author on
request, and its magnitude belongs here rather than only in those files, because making a file
available is not reporting. The analyser re-applies the
pre-amendment answer parser (first \texttt{ANSWER:} line) alongside the binding one (last line) to
every stored arm: 606 arm records in all. On 37 of them at least one item's prediction changes;
on 13 the arm's \emph{accuracy} changes. The rest of this paragraph is about that smaller count.
The largest accuracy move is 5 items of 120, on Phi-4's no-fault-plus-blind-inspector cell, and
the next is 4 items on Qwen3-14B's
grounded-inspector arm; the remaining eleven move by 2 items or fewer. The largest of these
exceeds the measured run-to-run jitter of 4 items, so on the arms where the scoring rule bites at
all it is a slightly larger source of movement than decoding noise. Both of the largest sit in the
defence arms, which are exploratory. Two confirmatory arms are among the thirteen and both move by
a single item: the primary backbone's no-fault dose cell and Phi-4's quarter-visibility occlusion
level. A one-item move is a quarter of the measured jitter and changes no verdict in this paper,
but it is a real sensitivity to the scoring rule and we print it rather than release it silently.
The per-arm difference for every arm is in the result files, available from the corresponding
author on request, so a reader can re-apply either
rule to the stored raw outputs.

\input{tables/T10-denominators}

\section{The two matched denominators, and the token premise}
\label{app:denominators}

Table~\ref{tab:denominators} exists because one metric name was carrying two quantities. The
moving-ceiling ratio $\prrm$ divides by the no-fault cell at the same $k$; the occlusion ratio
$\prrocc$ divides by one fixed ceiling, the fully re-grounded no-fault cell, at every visible
fraction. On Qwen3-14B the two zero points are 0.368 and 0.221 as ratios and 0.208 and 0.192 as
accuracies, so most of the apparent gap between the two families at zero is the denominator. The
first block prints both families in accuracy as well as in ratio for that reason.

The second block is the token premise the occlusion control rests on. Blanks preserve word count,
not tokenizer count, so the two ends of the slider are close but not equal in prompt length: the
fully masked arm carries 795.2, 782.5, 908.0 and 675.1 prompt tokens per item against 825.0,
865.8, 880.1 and 716.7 at full visibility, ratios of 0.964, 0.904, 1.032 and 0.942. On Phi-4 the
ratio exceeds 1, so the \emph{masked} arm is the longer one and still scores worse, a stronger
version of the control's argument than the one the paper needs.

The third block prints Family~B's ratio at all five visible fractions on all four backbones with
its paired intervals, so the endpoint contrast the kill rule reads can be checked against the
interior levels it is built from.

\input{tables/T4-cascade}

\section{Cross-family results}
\label{app:cascade}

Table~\ref{tab:cascade} reports the six cross-family tasks on all four backbones in one block of
24 rows: the deployed ratio at the strict pipeline, at the middle dose and at $k_{\max}$, and the
matched ratio at the last two. Every matched ceiling has landed: 48 re-grounded no-fault arms are
on disk, all 24 $k_{\max}$ cells carry a matched ratio, and the analyser records none absent.

Two columns a reader might look for are absent by construction, and the table's footnote says so
rather than leaving a gap. The deployed gain is the difference of two columns already printed, so
it is not repeated as a third. And $\prrm(k_{\min})$ equals $\prrd(k_{\min})$ exactly, because at
$k_{\min}$ the matched ceiling \emph{is} the strict no-fault arm, so printing it twice would
suggest two measurements where there is one.

Two readings of the table need care. First, the matched ratio equals the deployed one in 12 of 24
cells. That happens where the family's re-grounded no-fault arm reproduces the strict one, and in
this campaign it happens only on families already at or within one item of ceiling. Of the 48
arms, 22 reproduce the strict arm item for item; of the 12 equal cells, 11 share the ceiling
vector, and 1, Llama coin flip, has the same arm mean on 6 differing items and is a coincidence
rather than an unchanged ceiling. Llama-3.1-8B reproduces no ceiling at all and its
re-grounded ceilings move by up to 39 of 120 items: where there is room to move, the ceiling
moves, which is the argument for the moving denominator. Second, the injected fault inserts a
bogus step but does not hijack the answer line when the upstream text contains no number, so it is
weaker on the non-numeric families, and one fault across families is not claimed anywhere in this
paper.

\input{tables/T3-mechanism}

\begin{figure}[t]
\centering
\includegraphics[width=1.0\textwidth]{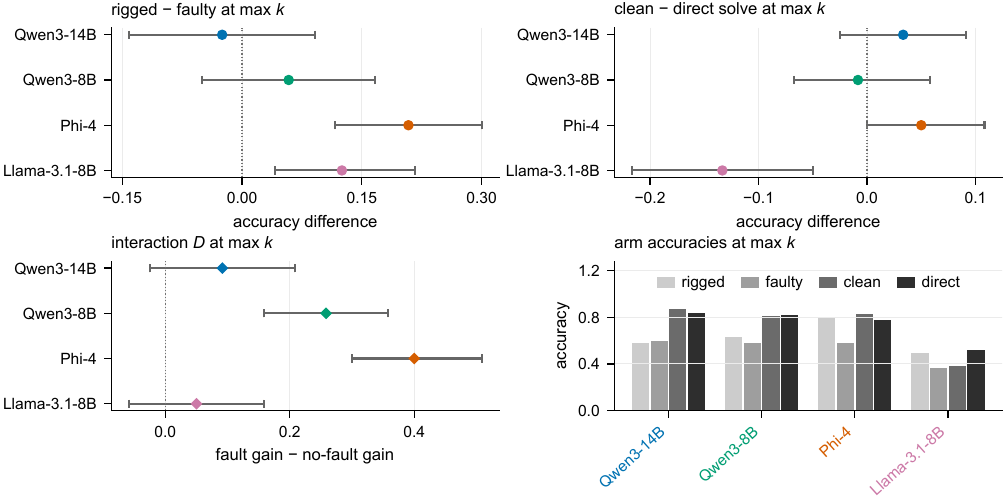}
\caption{Mechanism, per backbone, exploratory and two-sided. Points are paired differences with
95\% paired item bootstrap intervals; the dotted line is zero. \textbf{Top left}: replacing the
upstream message with a fixed uninformative line at $k_{\max}$. \textbf{Top right}: the fully
re-grounded no-fault pipeline against a single-agent direct solve. \textbf{Bottom left}: the
interaction $D(k_{\max})$ on the registered pipeline, the re-grounding gain under fault minus the
gain without it, which covers zero on Qwen3-14B and on Llama-3.1-8B; Figure~\ref{fig:pipelines}
carries the same estimand on both measured pipelines. \textbf{Bottom right}: the four arm
accuracies the contrasts are built from.}
\label{fig:mechanism}
\end{figure}

\section{The mechanism table, and the length-matched control}
\label{app:mechanism}

Table~\ref{tab:mechanism} carries the three mechanism contrasts, the four arm accuracies behind
them, the interaction at every dose, the smallest interaction this design detects, and the
post-pin length-matched rigging arm.

\paragraph{The interaction below $k_{\max}$.} The registration fixes $D$ at $k_{\max}$; the
interior doses are exploratory and are printed because an empty column invites the reader to
assume the worst. On Qwen3-14B $D(k{=}1) = +0.192$ $[+0.075, +0.308]$ excludes zero while
$D(k{=}2)$ and $D(k{=}3)$ do not, because the no-fault ceiling rises between $k = 1$ and $k = 2$.
On Qwen3-8B and Phi-4 every dose excludes zero; on Llama none does.

\paragraph{The length-matched arm, and what it changes.} The registered rigging control replaces
the upstream message with one short line, so it varies content and length together. The post-pin
arm replaces it with a digit-free filler of exactly the replaced message's character length. Its
accuracies are 0.408, 0.600, 0.825 and 0.425. Against the one-line control, the length-matched
filler is $-0.167$ $[-0.250, -0.083]$ on Qwen3-14B, the one backbone where length demonstrably
matters, and within the interval on the other three. Against the faulty arm it is $-0.192$
$[-0.308, -0.067]$ on Qwen3-14B, $+0.025$ $[-0.083, +0.133]$ on Qwen3-8B, $+0.242$ $[+0.150,
+0.333]$ on Phi-4 and $+0.058$ $[-0.033, +0.150]$ on Llama. The distraction account therefore
survives on exactly one backbone, Llama, whose anomaly disappears once the control is
length-matched. It is refuted on the primary backbone, where a longer content-free upstream costs
\emph{more} than the faulty one. And it is unsupported on Phi-4, where the anomaly survives at
$+0.242$.

\input{tables/T11-strict-direct}

\section{The strict pipeline against one call, and what the rigging interval supports}
\label{app:strict}

Table~\ref{tab:strict-direct} carries two blocks.

The first contrasts the \emph{strict} no-fault pipeline at $k = 0$, the thing a practitioner
would deploy, with a single direct call in the same answer format. The four-stage pipeline is
below one call on three backbones ($-0.267$, $-0.125$ and $-0.317$, all excluding zero) and
indistinguishable on the fourth ($+0.058$ $[+0.000, +0.125]$, McNemar $p = 0.118$; the interval's
lower bound sits exactly at zero, which is why the row is labelled indistinguishable rather than
above). Table~\ref{tab:mechanism} contrasts the fully \emph{re-grounded} no-fault cell against the
same direct solve, which is a different and kinder comparison; both are reported.

The second block says what the rigging control supports and no more. No equivalence margin is
registered anywhere in the analysis plan, so no two-one-sided-test procedure is run: a
margin chosen now would be a threshold set after seeing the interval it has to clear. A
non-significant contrast is therefore reported as a failure to detect together with the effect it
still admits. On Qwen3-14B it is not detected at $n = 120$, with the interval admitting
$\pm 0.117$; on Qwen3-8B it is not detected either, with $\pm 0.108$.

\input{tables/T13-defense-n120}

\section{The defence head-to-head, and the ceiling it is measured against}
\label{app:defense}

Table~\ref{tab:defense} carries the contrast against the closest prior work at the headline
standard: every arm on the same 120 items, on all four backbones, with per-arm Wilson intervals
and paired contrasts. Section~\ref{sec:mechanism} quotes the two contrasts; three further points
belong here.

First, the blind inspector is not a weak version of the grounded one, it is nothing. Its accuracy
is 0.208, 0.200, 0.175 and 0.058 against strict-pipeline values of 0.208, 0.200, 0.183 and 0.133.
At the per-item level the two arms are \emph{identical} on Qwen3-14B and Qwen3-8B, differing on 0
of 120 items, and differ on only 3 and 9 items on Phi-4 and Llama-3.1-8B. They are separate
measurements that converged rather than one file counted twice: the two arms issued 73 against 65
server calls on Qwen3-14B and 111 against 100 on Qwen3-8B. Two consequences follow. The
grounded-minus-blind and grounded-minus-strict contrasts are the \emph{same number} on the two
Qwen backbones and are not independent evidence there, which is why this appendix prints their
item counts side by side: $+73$ and $+73$, $+69$ and $+69$, $+73$ and $+72$, $+43$ and $+34$ of
120. And on Qwen3-14B the same identity holds without a fault, where \texttt{clean} and
\texttt{clean\_inspect} also differ on 0 of 120 items. That null belongs to one backbone. The
fourth block of Table~\ref{tab:defense} carries the same contrast on all four, as signed item
counts: 0 on Qwen3-14B, $+3$ in the inspector's favour on Qwen3-8B, and $-23$ and $-16$ against
it on Phi-4 and Llama-3.1-8B, the last two the only ones whose intervals exclude zero.
Section~\ref{sec:mechanism} gives those intervals. That block is filed apart from the rows above it because
every other contrast in this table starts from the faulted pipeline and this one does not, so
reading it as a fourth defence number would be wrong. Whatever a reviewer agent buys in this
setting, it does not buy it by reviewing.

Second, the two ceiling ratios in the second block are different quantities and we print both for
the reason Appendix~\ref{app:denominators} gives about the dose arms. Divided by the isomorphic
no-fault cell, the inspector pipeline with no fault at the same stage count, the grounded arm
reaches 0.980 $[0.931, 1.030]$, 0.979, 0.969 and 0.962. Divided by a no-fault cell at a different
stage count, the same arms read 1.441
$[1.247, 1.695]$, 1.120, 1.237 and 6.250 $[3.500, 16.667]$. The 6.250 is the clearest case in this
paper of a ratio describing its denominator rather than its numerator.

\input{tables/T8-pipelines}

\section{The two pipelines side by side}
\label{app:pipelines}

Table~\ref{tab:pipelines} reports pipeline~A (the registered four-stage extract / set up / compute
/ format decomposition) and pipeline~B (a three-stage understand / solve / answer decomposition)
on the same 120 items, under the same fault, with the same strict visibility rule and each with
its own matched ceilings. Each pipeline's interaction is taken at its own $k_{\max}$, which is 3
for A and 2 for B.

\paragraph{Where the registered pipeline starts, and what that does to $D$.} With no fault the
registered pipeline is the worst of the four configurations measured here on two backbones and
third of four on a third, against pipeline~B, full history and a single call. Re-grounding
repairs a handicap as readily as it repairs a fault, so that standing feeds straight into $D$.
Where the registered pipeline starts lowest, re-grounding buys the \emph{healthy} arm $+0.300$
and $+0.183$, leaving interactions of $+0.092$ and $+0.050$, the two smallest we report. On
Phi-4, where the registered pipeline is the best of the four, the no-fault gain is $-0.008$ and
$D$ is $+0.400$, the largest. This is the explanation for the paper's own weakest number, and it
is a statement about the configuration rather than about the backbone.

Pipeline~B is measured and not registered. It enters no confirmatory family, no kill clause reads
it, and every number from it is exploratory. It is reported because the registration's exclusion
rule is ``none post hoc'': a measured design arm on four backbones is either reported or its
exclusion is dated and justified, and no such row exists. The direction matters: including
pipeline~B strengthens the fault-specificity reading rather than weakening it.

\input{tables/T12-history-full}

\section{Full message history against predecessor-only visibility}
\label{app:history}

Table~\ref{tab:history} carries the external-validity condition: the same four-stage pipeline, the
same fault, the same 120 items, with every stage given the full message history instead of its
predecessor's message alone. Section~\ref{sec:history} quotes the contrasts; two readings need
care.

The first is what an indistinguishable contrast does and does not establish. On three backbones
the full-history arm differs from the predecessor-only arm at $k_{\max}$ by $+0.000$, $+0.008$ and
$-0.033$, a \emph{net} difference of $0$, $1$ and $4$ items of 120, with intervals that cover zero
and McNemar $p$ of 1.000, 1.000 and 0.219. A net is not an agreement, and here the two are far
apart: item by item the arms disagree on $6$, $7$ and $6$ items, so the $+0.000$ on Qwen3-14B is
three items won and three lost rather than an identical vector, and on Llama-3.1-8B, which the
sentence above excludes, the disagreement is $13$ items for a net of $-0.075$. That churn exceeds
this campaign's measured run-to-run jitter of 4 items on all four backbones, which is what two
jobs run three days apart should be expected to produce, and it is why the claim made here is the
weak one. The contrast is a failure to detect a difference at $n = 120$, not a
demonstration that none exists, and we report it with the same discipline as the rigging control:
the widest of those intervals still admits $\pm 0.075$. What it does rule out is the strong form
of the external-validity objection, that the dose response is an artefact of information scarcity,
because an arm with strictly more information available at every stage reproduces it.

The second is Llama, where full history is measurably \emph{worse}: $-0.075$ $[-0.133, -0.017]$,
$p = 0.022$, where the $-0.133$ is this interval's lower bound and not a point estimate; three
other tables print that same value as one. This is the thinnest margin the paper claims, so we state it against the
paper's own yardstick: 9 items of 120, against a within-campaign run-to-run jitter of 4 items,
which the analysis plan requires any claimed conclusion to clear. It clears it by a factor of
about two, where every other contrast this paper claims clears it by three or more. On the one
backbone in this set with a low ceiling, handing every stage more context costs accuracy rather
than buying it, which is consistent with the lost-in-the-middle literature
\citep{liu2024lostmiddle} and inconsistent with any claim that more visibility is uniformly
better. It is also a second reason, beside the absent fault cost, to treat that backbone as a
boundary case rather than a fourth replication.

The arm is exploratory and two-sided: it is not in the registered plan, and its direction was not
written down before the data.

\input{tables/T9-cost-headline}

\section{Cost at the headline standard}
\label{app:cost}

Table~\ref{tab:cost-headline} is the cost curve at the headline standard: 120 items per arm on
four backbones, both named ratios, paired bootstrap intervals, and average tokens defined as
prompt plus completion per item read from the recorded server usage. The first re-grounded stage
buys $+0.394$ of matched retention for $+59.8$ tokens, and the stages after it buy nothing at any
price.

\input{tables/T6-contribution-three}

\section{Topology search as a negative control}
\label{app:contribution-three}

Table~\ref{tab:contribution-three} carries both halves of the topology-search arm: the prospective
evolutionary search scored on a held-out split, and the retrospective front computed over the
whole bank. The contrasts the paper reports are the matched ones. Prospectively, ours minus
accuracy-only is $-0.5$ points on the held-out split, matched by construction because both arms
are scored on the same split with the same scoring. Random selection scores 0.991 on the same
split, above both search arms: ours minus random is $-0.6$ points and accuracy-only minus random
is $-0.1$ points, both unmatched because the fronts hold 3, 3 and 5 points. Retrospectively the
two fronts differ in cost and in clean accuracy, so their apparent advantage of $+0.7$ points is
unmatched and is not a result; at the matched cost tier both strategies select the same topology
and the gain is $+0.0$ points. Nothing from this arm is presented as a contribution, and the arm
it loses to is named rather than omitted.

\input{tables/T5-flat-regime}

\begin{figure}[t]
\centering
\includegraphics[width=1.0\textwidth]{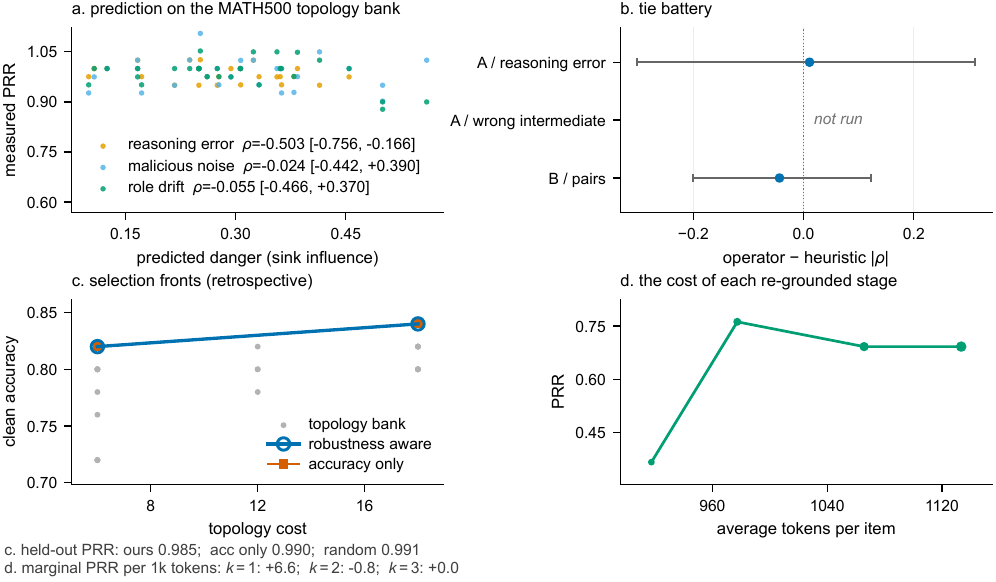}
\caption{The flat regime on MATH500 and the cost of each re-grounded stage; primary backbone
only, exploratory throughout. \textbf{(a)} Each point is one topology: predicted danger from the structural
operator against measured retention, per fault family. \textbf{(b)} The tie battery: the
operator's rank correlation minus the one-line heuristic's, with paired bootstrap intervals; the
cell whose fault matches the headline campaign has not run, and the bank injects three other fault
families. \textbf{(c)} Retrospective selection fronts. \textbf{(d)} The cost of each re-grounded stage on
the same 120 items as the dose ladder, tokens counted as prompt plus completion per item and
markers growing with $k$, so the smallest is $k=0$; the headline cost curve is Figure~\ref{fig:pipelines}c. Panels (c) and (d) carry their
numbers on a line beneath the panel rather than inside it.}
\label{fig:flat}
\end{figure}
\begin{figure}[t]
\centering
\includegraphics[width=1.0\textwidth]{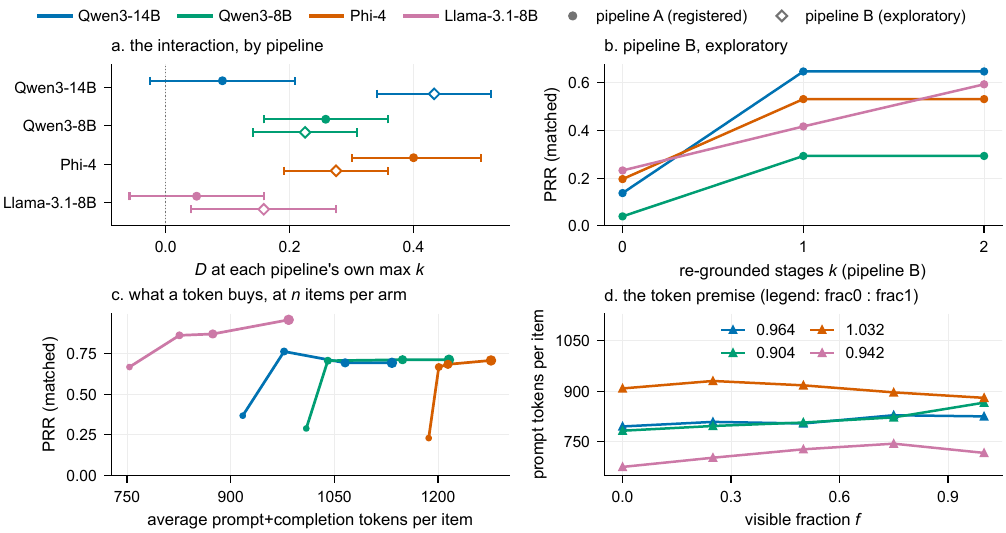}
\caption{The second pipeline, the cost curve at the headline standard, and the occlusion token
premise. \textbf{(a)} The fault-by-re-grounding interaction $D$ under each pipeline at its own
$k_{\max}$, with 95\% paired item bootstrap intervals: it excludes zero on two of four backbones
under the registered pipeline~A (filled circle) and on four of four under pipeline~B (open
diamond), which is measured and \emph{not} registered, so every number it produces is
exploratory. \textbf{(b)} Pipeline~B's matched dose curve. \textbf{(c)} Matched retention against
average prompt and completion tokens, markers growing with $k$. \textbf{(d)} Occlusion prompt
tokens by visible fraction; on Phi-4 the masked arm is longer.}
\label{fig:pipelines}
\end{figure}

\section{Levers that are not levers}
\label{app:nonlevers}

Two interventions were measured as candidate second levers and neither pays. Both are exploratory
and both are reported here in full rather than summarised, because a null is only useful at the
size it was measured.

Redundancy does not reproduce as a second lever: under the re-run fault the blind aggregator
already retains 1.000, 1.022 and 1.011 at one replica on the first three backbones and 0.538 on
Llama, against grounded values of 0.980, 1.000, 1.000 and 0.717. Robustness-aware topology search
does not pay either. On a MATH500 bank of 30
topologies it returns three Pareto picks at held-out retention 0.985, against 0.990 for
accuracy-only search and 0.991 for \emph{random} selection; ours minus accuracy-only is $-0.5$
points and ours minus random $-0.6$, so random is the best arm here. At the matched cost tier both
strategies pick the same topology, gaining $+0.0$ points, and a calibrated propagation operator
does not outrank a one-line heuristic under the bank's reasoning-error fault ($+0.011$ $[-0.302,
+0.312]$). The bank sits near a plateau and the cell matching our own fault has $n = 0$, so this
is a null on one bank on one backbone (Appendix~\ref{app:flat}).

\section{The flat regime in full}
\label{app:flat}

Table~\ref{tab:flat} carries four blocks that report different quantities, which is why each block
names its own columns: the structural predictor's rank correlation with measured retention per
fault family, the coverage of the registered standard benchmarks, the tie battery between the
calibrated operator and the one-line heuristic, and the per-topology subset ranking. Everything in
this section is exploratory, single-backbone and single-bank, and the main text draws no
architectural conclusion from it.

\paragraph{What the bank can and cannot say.} On 30 topologies scored on 50 items each, a
structural predictor derived from a resolvent propagation operator in the lineage of
\citet{katz1953status}, \citet{bonacich1987power} and \citet{leontief1936quantitative} correlates
with measured retention for one fault family and for neither of the other two: Spearman
$\rho = -0.503$ $[-0.756, -0.166]$ for a reasoning-error fault, $-0.024$ $[-0.442, +0.390]$ for
injected noise and $-0.055$ $[-0.466, +0.370]$ for role drift (Figure~\ref{fig:flat}a). Three
limits bound every reading of this bank. Clean accuracy across it runs from 0.720 to 0.840 and
retention under the reasoning-error fault from 0.750 to 1.026, so the bank is close to a plateau
and there is little variance for a predictor to explain. The one informative correlation is under
a fault family this paper does not inject. And the cell whose fault family matches the headline
campaign has $n = 0$: it was never run. A null measured here is a statement about this bank under
these faults, and nothing in it licenses a claim that pipeline topology does not matter.

\paragraph{The four entries that cannot move.} Four of the bank's entries are single-round and
acyclic, and they score exactly 1.000 under every fault because the injected fault reaches no
downstream stage at all. We report the bank statistics over all 30 entries with no exclusion, and
we state the reason rather than leaving it to a listing: those four contribute no variance rather
than favourable variance, so dropping them would move every bank-level number in the direction
that flatters the paper's own reading. The listing, with those four marked, is available with the
code from the corresponding author on request.

\paragraph{A calibrated operator against a one-line heuristic.} Against a proximity-to-sink
heuristic that fits on one line, the calibrated operator's rank correlation with retention is
0.503 against 0.492 on the first cell, a difference of $+0.011$ $[-0.302, +0.312]$ over 30
topologies; pooled over 90 pairs it is 0.198 against 0.242, $-0.044$ $[-0.200, +0.122]$; on eight
multi-fault topologies the convex form scores 0.678 against 0.740 for the structural one,
$-0.062$, without an interval (Figure~\ref{fig:flat}b). The intervals are wide, so the honest
reading is that the two are indistinguishable in our tests, not that they are equivalent.

\begin{figure}[t]
\centering
\includegraphics[width=1.0\textwidth]{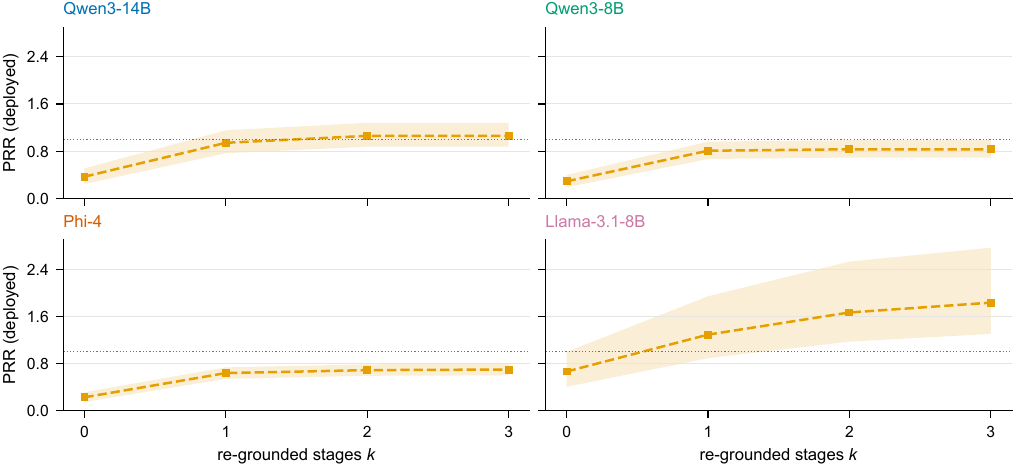}
\caption{The deployed ratio, one shared vertical axis across all four panels. $\prrd$ divides by
the deployed strict pipeline at $k = 0$, a ceiling that does not move with $k$; the analysis plan
calls it an operational number and never a restoration. Bands are 95\% paired item bootstraps. The
dotted line is the unit ratio: above it the faulty arm scores above the strict no-fault pipeline,
which is a statement about the denominator and not about the fault. This panel set is in the
appendix, and the matched ratio that every claim rests on is Figure~\ref{fig:dose}, so the two are
never read off one axis.}
\label{fig:deployed}
\end{figure}

\section{The deployed ratio}
\label{app:deployed}

Figure~\ref{fig:deployed} carries $\prrd$ on all four backbones on one shared axis. Drawing it
beside the matched ratio would need the Llama panel scaled to 2.5 while the other three ran to
1.25, destroying cross-panel comparability for the one quantity the paper calls operational; it
is separated here for that reason. Nine deployed ratios
across the three cross-family doses exceed 1.0; the matched column resolves seven of them, and the
remainder sit at Phi-4 boolean evaluation, where both ratios read 1.008 because the family's
strict ceiling is itself 0.992 and one item separates the arms.

\section{Reproducibility}
\label{app:repro}

Available from the corresponding author on request: the pipeline and its stage prompts for both pipelines; the fault
injector; the re-grounding and occlusion switches; every experiment driver; the analyser with its
self-test suite; the lint that binds each registered sentence to the analyser function
implementing it; the registration document and its dated deviations; and the per-item correctness
vector, parsed prediction, raw final-stage output and token counts for every arm of every
backbone. Every table and figure in this paper is generated from those result files by a single
regeneration command. A claims-trace check verifies every numeral printed in the manuscript against a recorded result file.

\paragraph{What the campaign cost, in the units the result files record.}
We report inference rather than wall clock, because wall clock lives in the scheduler's accounting
and in no result file, while every file records its own model-call count and the split between
fresh samples and prompt-cache replays. Across the 65 delivered result files and 1042 arms the
campaign spent 570090 model calls: 64478 fresh temperature-zero samples and 505612 deterministic
replays from the shared prompt cache, with 0 call errors and 0 empty completions. Per backbone the
calls are 350280, 79030, 70390 and 70390, and the primary carries the larger share because the
replica lane and the twenty-seed cell both run on it. A replay is not a saving hidden from the
reader: at temperature zero a cache hit returns the same completion the same prompt produced
before, which is why Appendix~\ref{app:replay} reports the replay status of every arm separately
rather than folding it into a total.
A further 15 files carrying 161280 calls are \emph{excluded} from that budget, and naming them
matters more than the arithmetic. They are every \texttt{cascade\_suite.json},
\texttt{occlusion.json}, \texttt{redundancy.json} and \texttt{redundancy\_blind.json}: files from
the older writer, whose channel counters do not exist, and which are exactly the superseded arms
this paper does not read. The CASCADE and redundancy arms were re-run under \wi{} into
\texttt{.wrong\_intermediate.json} siblings and the first occlusion build is disowned by its own
runner. Counting them would inflate the disclosed budget by work no delivered number rests on.

\paragraph{A clamp in a superseded plotting helper.} One plotting helper applied an asymmetric
clamp: a ceiling of
1.00 on the baseline series against 1.05 on the treatment series, which flatters the treatment by
construction. It read a superseded result tree, it was never invoked by the regeneration driver,
and nothing in the delivered figure path imports it; the script that draws every panel in this
paper contains no clip and no clamp. The helper and five sibling scripts from the same superseded
tree have been deleted, with their hashes recorded in the regeneration driver so that the deletion
is auditable rather than merely asserted.

\paragraph{What the code does and does not rebuild.} One command rebuilds every table and
figure in this paper from the result files, and the regeneration driver names the
delivered analysis explicitly, so the rebuild path and the delivered path are the same path. It
does not rebuild the result files from the cluster: that needs the inference server and the GPU
allocation, and the command that prints the exact submission invocations is included instead.
Inference ran on an academic GPU cluster behind a local paged-attention server
\citep{kwon2023vllm}; no hosted model API was used. Code and per-run logs are available from the
corresponding author on request.

%% file: tables/T2-families.tex
\begin{table}[t]
\centering
\footnotesize
\caption{Registered families and the kill rule. Families A and B are the confirmatory tests (McNemar exact one-sided, Holm at the registered family size). The kill rule is a point-estimate rule and fires only when a clause fails on at least the required number of backbones. Breakdown margins report the smallest margin at which a clause would fire, which is the informative quantity rather than a fixed band around the registered one. The blocks are, in order: Family A: endpoint k0 vs kmax, McNemar exact one-sided, Holm over the registered m=4 at 0.05; Family B: occlusion frac=0 vs frac=1, McNemar exact one-sided, Holm over the registered m=4 at 0.05; kill rule: a clause fires only on at least the required number of backbones.}
\label{tab:families}
\begin{tabular}{@{}l p{0.32\linewidth} r r p{0.25\linewidth}@{}}
\toprule
\midrule
\multicolumn{5}{@{}l}{\emph{Family A: endpoint k0 vs kmax, McNemar exact one-sided, Holm over the registered m=4 at 0.05}} \\
quantity & backbone & Holm-adjusted $p$ & unadjusted $p$ & decision \\
endpoint contrast & Qwen3-14B & 8.3e-12 & 2.8e-12 & reject \\
 & Qwen3-8B & 1.1e-13 & 2.8e-14 & reject \\
 & Phi-4 & 8.3e-12 & 2.8e-12 & reject \\
 & Llama-3.1-8B & 2.1e-6 & 2.1e-6 & reject \\
decision & backbones holding & 4 & 4 & holds on every backbone (4/4) \\
\midrule
\multicolumn{5}{@{}l}{\emph{Family B: occlusion frac=0 vs frac=1, McNemar exact one-sided, Holm over the registered m=4 at 0.05}} \\
quantity & backbone & Holm-adjusted $p$ & unadjusted $p$ & decision \\
endpoint contrast & Qwen3-14B & 6.4e-13 & 1.6e-13 & reject \\
 & Qwen3-8B & 6.5e-12 & 2.2e-12 & reject \\
 & Phi-4 & 1.1e-11 & 5.5e-12 & reject \\
 & Llama-3.1-8B & 3.7e-6 & 3.7e-6 & reject \\
decision & backbones holding & 4 & 4 & holds on every backbone (4/4) \\
\midrule
\multicolumn{5}{@{}l}{\emph{kill rule: a clause fires only on at least the required number of backbones}} \\
clause & rule & backbones failing & required & verdict \\
1 PRR\_deployed gain & PRR\_deployed(k3) - PRR\_deployed(k0) \mbox{$<$ 0.20} & 0 & 2 & does not fire \\
2 PRR\_matched gain & PRR\_matched(k3) - PRR\_matched(k0) \mbox{$<$ 0.20} & 0 & 2 & does not fire \\
3 adjacent decrease (interval rule) & an adjacent decrease in acc(F,k), k=0..3, whose item-paired one-sided 95\% lower bound excludes zero & 0 & 2 & does not fire \\
4 occlusion endpoint & occlusion PRR\_matched(frac=0) \mbox{$\geq$ PRR\_matched(frac=1)} - 0.10 & 0 & 2 & does not fire \\
kill rule & clauses firing / backbones & 0 & 4 & not killed: no clause fires on \mbox{$\geq$ 2} of 4 backbones \\
\bottomrule
\end{tabular}
\vspace{0.4em}
\begin{minipage}{\linewidth}\footnotesize
Families A and B share the max $k$ measurement by construction: occlusion at frac${=}1$ and the dose arm at max $k$ are the same calls, with zero item disagreement on every backbone. They are therefore not independent evidence about that arm, and the paper says so rather than counting them twice.
\end{minipage}
\end{table}

%% file: tables/T2b-kill-rule-behaviour.tex
\begin{table}[t]
\centering
\footnotesize
\caption{How the registered kill rule behaves: the smallest margin at which each clause would fire, the transfer threshold and the false-kill rates, and the rule's power against a simulated true gain. The clauses themselves, and the confirmatory families they guard, are in Table~\ref{tab:families}. Breakdown margins report the smallest margin at which a clause would fire, which is the informative quantity rather than a fixed band around the registered one. The blocks are, in order: breakdown margins: the smallest margin at which the clause would fire; transfer, and the adjacent clause's false-kill rate under a global plateau; the kill rule's POWER: how often clause 1 fires when the true gain is small (simulated).}
\label{tab:kill-rule-behaviour}
\begin{tabular}{@{}l p{0.36\linewidth} r r p{0.27\linewidth}@{}}
\toprule
\multicolumn{5}{@{}l}{\emph{breakdown margins: the smallest margin at which the clause would fire}} \\
quantity & definition & breakdown & registered & where \\
PRR (deployed) gain & second-smallest gain PRR(kmax)-PRR(k0): \mbox{$\geq$ 2} backbones fail for any kill margin above it & 0.542 & 0.200 & at Qwen3-8B, +28.4 items from the rule \\
PRR (matched) gain & second-smallest gain PRR(kmax)-PRR(k0): \mbox{$\geq$ 2} backbones fail for any kill margin above it & 0.325 & 0.200 & at Qwen3-14B, +13.0 items from the rule \\
occlusion endpoint & a backbone's occlusion kill fires at every margin \mbox{$\geq$ its} gap PRR(frac1)-PRR(frac0); the registered rule needs \mbox{$\geq$ 2} backbones & 0.471 & 0.100 & at Qwen3-14B \\
adjacent decrease & largest one-sided 95\% lower bound (items) of any adjacent decrease: that pair counts at every threshold below it & +0.0 & 5 & at Qwen3-8B \\
run-to-run jitter & items at $T{=}0$, IMPORTED from prior jobs & 2 & -- & a conclusion inside the jitter is not claimed \\
run-to-run jitter & items, MEASURED in this campaign & 4 & -- & largest per-arm item disagreement against the headline file: 13 flips across 8 arms \\
\midrule
\multicolumn{5}{@{}l}{\emph{transfer, and the adjacent clause's false-kill rate under a global plateau}} \\
quantity & definition & measured & registered & verdict \\
transfer threshold & third-largest PRR\_matched(kmax) among backbones with PRR\_matched(k0) \mbox{$<$ 0.5:} 3 of 4 pass at every threshold up to it & 0.692 & 0.900 & target not met; reported as measured \\
eligible backbones & PRR (matched) at min $k$ $<0.500$ & 3 & 4 & the threshold is the third-largest among the eligible \\
false-kill rate & per backbone, against $\alpha$ & 0.143 & 0.050 & closed form, no rate typed by hand \\
false-kill rate & the family rule as a whole & 0.100 & 0.050 & over 12 unadjusted one-sided pairs \\
\midrule
\multicolumn{5}{@{}l}{\emph{the kill rule's POWER: how often clause 1 fires when the true gain is small (simulated)}} \\
quantity & definition & power & true gain & note \\
clause 1 power & the rule fires when the true gain is zero & 0.999 & 0.000 & a pipeline with no dose response at all \\
 & the rule fires at the registered margin itself & 0.705 & 0.200 & a boundary case: about half the time, as it should be \\
 & the rule fires at the gains this campaign measured & 0.000 & as measured & the rule had every chance to fire here and did not \\
detection bound & largest true gain still detected at 80\% power & 0.150 & -- & above this the rule stops being able to fire \\
\bottomrule
\end{tabular}
\vspace{0.4em}
\begin{minipage}{\linewidth}\footnotesize
The clauses these margins belong to, and the confirmatory families they guard, are in Table~\ref{tab:families}.
\\ \emph{How the power block above was simulated.} Clause 1 of the registered kill rule fires on a backbone when PRR\_deployed(k\_max) - PRR\_deployed(k\_0) \mbox{$<$ the} registered margin. Its sampling distribution is the paired-item bootstrap of that gain on the LANDED vectors of each backbone, RECENTRED on a candidate true gain g -- so the dispersion is the campaign's own and only the location is hypothetical. One draw per backbone per simulation, backbones independent; the rule fires when \mbox{$\geq$ the} registered number of backbones draw a gain below the margin. Power at g is the firing rate over the simulated draws. The curve is read at g = the margin (where the rule should fire about half the time) and below it (where a paper that deserves to be killed lives).
\end{minipage}
\end{table}

%% file: tables/T10-denominators.tex
\begin{table}[t]
\centering
\footnotesize
\caption{The two matched denominators, and the token premise of the occlusion control. PRR (matched) (Family~A) divides by the moving matched ceiling: the no-fault cell at the same $k$ (\texttt{clean\_k} at that $k$); PRR (occ) (Family~B, the occlusion slider) divides by the fixed ceiling: the fully re-grounded no-fault cell (\texttt{clean} in the occlusion file), one denominator at every visible fraction. They are different quantities and are named differently everywhere they appear. Both are printed in accuracy as well as in ratio, because the gap between the two zero points is mostly a denominator difference and that is invisible in ratios alone. The second block is the occlusion prompt-token count per level: masking replaces each word with a fixed token, so the two arms are close in length, and where the ratio exceeds 1 the \emph{masked} arm is the longer one and still scores worse. The blocks are, in order: the two zero points: same paper, two denominators; occlusion prompt tokens per item, by visible fraction $f$; Family B: PRR (occ) by visible fraction $f$, with paired item bootstrap intervals.}
\label{tab:denominators}
\begin{tabular}{llrrrrrl}
\toprule
\multicolumn{8}{@{}l}{\emph{the two zero points: same paper, two denominators}} \\
backbone & family & zero-point arm & acc. & ceiling arm & ceiling acc. & ratio & ratio name \\
Qwen3-14B & A (dose) & reground\_k0 & 0.208 & clean (dose file, strict) & 0.567 & 0.368 & PRR (matched) at $k{=}0$ \\
 & B (occlusion) & frac\_0.0 & 0.192 & clean (occlusion file, re-grounded) & 0.867 & 0.221 & PRR (occ)$(f{=}0)$ \\
Qwen3-8B & A (dose) & reground\_k0 & 0.200 & clean (dose file, strict) & 0.692 & 0.289 & PRR (matched) at $k{=}0$ \\
 & B (occlusion) & frac\_0.0 & 0.200 & clean (occlusion file, re-grounded) & 0.808 & 0.247 & PRR (occ)$(f{=}0)$ \\
Phi-4 & A (dose) & reground\_k0 & 0.192 & clean (dose file, strict) & 0.833 & 0.230 & PRR (matched) at $k{=}0$ \\
 & B (occlusion) & frac\_0.0 & 0.183 & clean (occlusion file, re-grounded) & 0.825 & 0.222 & PRR (occ)$(f{=}0)$ \\
Llama-3.1-8B & A (dose) & reground\_k0 & 0.133 & clean (dose file, strict) & 0.200 & 0.667 & PRR (matched) at $k{=}0$ \\
 & B (occlusion) & frac\_0.0 & 0.142 & clean (occlusion file, re-grounded) & 0.383 & 0.370 & PRR (occ)$(f{=}0)$ \\
\midrule
\multicolumn{8}{@{}l}{\emph{occlusion prompt tokens per item, by visible fraction $f$}} \\
backbone & $f{=}0.000$ & $f{=}0.250$ & $f{=}0.500$ & $f{=}0.750$ & $f{=}1.000$ & $f{=}0$ : $f{=}1$ & longer arm \\
Qwen3-14B & 795.2 & 808.7 & 804.0 & 828.5 & 825.0 & 0.964 & visible \\
Qwen3-8B & 782.5 & 796.6 & 806.8 & 822.1 & 865.8 & 0.904 & visible \\
Phi-4 & 908.0 & 930.1 & 917.1 & 896.3 & 880.1 & 1.032 & masked \\
Llama-3.1-8B & 675.1 & 702.4 & 727.4 & 744.1 & 716.7 & 0.942 & visible \\
\midrule
\multicolumn{8}{@{}l}{\emph{Family B: PRR (occ) by visible fraction $f$, with paired item bootstrap intervals}} \\
backbone & quantity & $f{=}0.000$ & $f{=}0.250$ & $f{=}0.500$ & $f{=}0.750$ & $f{=}1.000$ &  \\
Qwen3-14B & ratio & 0.221 & 0.269 & 0.433 & 0.558 & 0.692 &  \\
 & [95\%] & [0.144, 0.304] & [0.186, 0.359] & [0.340, 0.528] & [0.459, 0.657] & [0.598, 0.781] &  \\
Qwen3-8B & ratio & 0.247 & 0.258 & 0.474 & 0.557 & 0.711 &  \\
 & [95\%] & [0.161, 0.340] & [0.172, 0.349] & [0.367, 0.584] & [0.455, 0.660] & [0.606, 0.814] &  \\
Phi-4 & ratio & 0.222 & 0.384 & 0.444 & 0.576 & 0.707 &  \\
 & [95\%] & [0.141, 0.310] & [0.287, 0.489] & [0.340, 0.553] & [0.469, 0.686] & [0.608, 0.806] &  \\
Llama-3.1-8B & ratio & 0.370 & 0.370 & 0.413 & 0.587 & 0.957 &  \\
 & [95\%] & [0.224, 0.538] & [0.225, 0.533] & [0.256, 0.596] & [0.415, 0.773] & [0.760, 1.204] &  \\
\bottomrule
\end{tabular}
\vspace{0.4em}
\begin{minipage}{\linewidth}\footnotesize
Prompt tokens are read from the recorded \texttt{usage}; the mask holds word count, not tokenizer count, so the two arms are close but not equal in length. Accuracies in the first block are the arm means behind the ratios beside them.
\end{minipage}
\end{table}

%% file: tables/T4-cascade.tex
\begin{table}[t]
\centering
\footnotesize
\caption{PRR (deployed) divides by the family's own no-fault arm at the strict pipeline; PRR (matched) divides by the moving matched ceiling: the no-fault cell at the same $k$ (\texttt{clean\_k} at that $k$). Every matched ceiling has landed: 48 re-grounded no-fault arms are on disk, 24 max $k$ cells carry a matched ratio, and 0 is recorded absent. The matched ratio equals the deployed one in 12 of 24 cells. That happens where the family's re-grounded ceiling is at, or within one item of, its strict ceiling: 22 of the 48 arms reproduce the strict arm item for item, and of the 12 equal cells 11 share the ceiling vector while 1 has the same mean on differing items and is a coincidence rather than an unchanged ceiling. Llama-3.1-8B reproduces no ceiling at all, and its re-grounded ceilings move by up to 39 of 120 items: where there is room to move, the ceiling moves, which is why a fixed denominator would be the wrong one. On non-numeric families the injected fault inserts one bogus step and does not hijack the answer line, so it is weaker there and one fault across families is not claimed.}
\label{tab:cascade}
\begin{tabular}{llrlllll}
\toprule
 &  &  & \multicolumn{3}{c}{PRR (deployed)} & \multicolumn{2}{c}{PRR (matched)} \\
\cmidrule(lr){4-6}\cmidrule(lr){7-8}
family & backbone & clean & min $k$ & $k{=}1$ & max $k$ & $k{=}1$ & max $k$ \\
\midrule
bool eval & Qwen3-14B & 1.000 & 0.975 [0.942, 1.000] & 1.000 [1.000, 1.000] & 1.000 [1.000, 1.000] & 1.000 [1.000, 1.000] & 1.000 [1.000, 1.000] \\
 & Qwen3-8B & 1.000 & 0.967 [0.933, 0.992] & 1.000 [1.000, 1.000] & 1.000 [1.000, 1.000] & 1.000 [1.000, 1.000] & 1.000 [1.000, 1.000] \\
 & Phi-4 & 0.992 & 0.832 [0.758, 0.900] & 1.008 [1.000, 1.026] & 1.008 [1.000, 1.026] & 1.008 [1.000, 1.026] & 1.008 [1.000, 1.026] \\
 & Llama-3.1-8B & 0.958 & 1.009 [0.966, 1.054] & 1.017 [0.974, 1.072] & 1.035 [1.000, 1.082] & 0.975 [0.942, 1.000] & 0.992 [0.975, 1.000] \\
\midrule
coin flip & Qwen3-14B & 0.992 & 0.992 [0.975, 1.000] & 1.000 [0.975, 1.026] & 1.000 [0.975, 1.026] & 0.992 [0.975, 1.000] & 0.992 [0.975, 1.000] \\
 & Qwen3-8B & 0.992 & 0.916 [0.864, 0.966] & 1.008 [1.000, 1.026] & 1.008 [1.000, 1.026] & 1.000 [1.000, 1.000] & 1.000 [1.000, 1.000] \\
 & Phi-4 & 1.000 & 0.733 [0.650, 0.808] & 0.850 [0.783, 0.908] & 0.900 [0.842, 0.950] & 0.850 [0.783, 0.908] & 0.900 [0.842, 0.950] \\
 & Llama-3.1-8B & 0.975 & 0.940 [0.881, 1.000] & 1.000 [0.958, 1.044] & 1.000 [0.958, 1.044] & 1.000 [0.958, 1.044] & 1.000 [0.958, 1.044] \\
\midrule
grid nav & Qwen3-14B & 1.000 & 0.025 [0.000, 0.058] & 0.225 [0.150, 0.300] & 0.225 [0.150, 0.300] & 0.225 [0.150, 0.300] & 0.225 [0.150, 0.300] \\
 & Qwen3-8B & 0.933 & 0.054 [0.017, 0.099] & 0.062 [0.025, 0.112] & 0.089 [0.043, 0.145] & 0.059 [0.025, 0.103] & 0.084 [0.042, 0.136] \\
 & Phi-4 & 1.000 & 0.033 [0.008, 0.067] & 0.058 [0.017, 0.100] & 0.058 [0.017, 0.100] & 0.058 [0.017, 0.100] & 0.058 [0.017, 0.100] \\
 & Llama-3.1-8B & 0.950 & 0.254 [0.177, 0.336] & 0.518 [0.420, 0.619] & 0.561 [0.462, 0.661] & 0.492 [0.400, 0.583] & 0.533 [0.442, 0.625] \\
\midrule
kinship & Qwen3-14B & 0.958 & 0.896 [0.826, 0.964] & 0.957 [0.890, 1.018] & 0.957 [0.890, 1.018] & 0.932 [0.874, 0.983] & 0.932 [0.874, 0.983] \\
 & Qwen3-8B & 0.383 & 0.630 [0.446, 0.852] & 0.609 [0.429, 0.821] & 0.565 [0.400, 0.758] & 0.718 [0.489, 1.000] & 0.667 [0.455, 0.938] \\
 & Phi-4 & 0.792 & 0.621 [0.515, 0.728] & 0.589 [0.483, 0.701] & 0.695 [0.588, 0.804] & 0.554 [0.452, 0.663] & 0.653 [0.549, 0.761] \\
 & Llama-3.1-8B & 0.567 & 0.897 [0.735, 1.083] & 0.912 [0.750, 1.105] & 0.912 [0.750, 1.105] & 0.954 [0.785, 1.150] & 0.954 [0.785, 1.150] \\
\midrule
logic deduction & Qwen3-14B & 1.000 & 0.775 [0.700, 0.850] & 0.958 [0.917, 0.992] & 0.975 [0.942, 1.000] & 0.958 [0.917, 0.992] & 0.975 [0.942, 1.000] \\
 & Qwen3-8B & 1.000 & 0.883 [0.825, 0.933] & 0.900 [0.842, 0.950] & 0.958 [0.917, 0.992] & 0.900 [0.842, 0.950] & 0.958 [0.917, 0.992] \\
 & Phi-4 & 1.000 & 0.300 [0.217, 0.383] & 0.758 [0.683, 0.833] & 0.925 [0.875, 0.967] & 0.758 [0.683, 0.833] & 0.925 [0.875, 0.967] \\
 & Llama-3.1-8B & 0.825 & 0.283 [0.196, 0.376] & 0.606 [0.490, 0.731] & 0.788 [0.667, 0.923] & 0.545 [0.447, 0.643] & 0.729 [0.626, 0.833] \\
\midrule
object swap & Qwen3-14B & 1.000 & 0.917 [0.867, 0.958] & 1.000 [1.000, 1.000] & 1.000 [1.000, 1.000] & 1.000 [1.000, 1.000] & 1.000 [1.000, 1.000] \\
 & Qwen3-8B & 0.933 & 0.804 [0.730, 0.874] & 0.946 [0.876, 1.019] & 0.955 [0.882, 1.029] & 0.883 [0.825, 0.933] & 0.892 [0.833, 0.942] \\
 & Phi-4 & 1.000 & 0.858 [0.792, 0.917] & 0.983 [0.958, 1.000] & 0.983 [0.958, 1.000] & 0.983 [0.958, 1.000] & 0.983 [0.958, 1.000] \\
 & Llama-3.1-8B & 0.625 & 0.947 [0.816, 1.097] & 1.107 [0.963, 1.279] & 1.133 [0.988, 1.308] & 0.954 [0.831, 1.093] & 0.944 [0.824, 1.081] \\
\bottomrule
\end{tabular}
\vspace{0.4em}
\begin{minipage}{\linewidth}\footnotesize
Intervals are $95\%$ paired item bootstraps. No cell in the matched column is pending. The deployed gain PRR (deployed) at max $k$ minus PRR (deployed)at min $k$ is the difference of two columns printed here and is not repeated as a third. PRR (matched) at min $k$ equals PRR (deployed) at min $k$ by construction -- at min $k$ the matched ceiling IS the strict \texttt{clean} arm -- so it is not printed twice either.
\end{minipage}
\end{table}

%% file: tables/T3-mechanism.tex
\begin{table}[t]
\centering
\footnotesize
\caption{Mechanism, per backbone (exploratory, outside the corrected families). The rigging control replaces the upstream message with a fixed uninformative line; the reference row is a single-agent direct solve. $D$ is the fault by re-grounding interaction. Intervals are paired item bootstraps. The blocks are, in order: component accuracies behind the contrasts above; the interaction at every dose (exploratory: the registration fixes $D$ at max $k$); the length-matched rigging control (exploratory, post-pin): a content-free upstream of the SAME length.}
\label{tab:mechanism}
\begin{tabular}{lllllll}
\toprule
backbone & rig $-$ fault at max $k$ & clean $-$ direct at max $k$ & $D$ at max $k$ & detectable $D$ \\
\midrule
Qwen3-14B & -0.025 [-0.142, +0.092] & +0.033 [-0.025, +0.092] & +0.092 [-0.025, +0.208] & 0.180 \\
Qwen3-8B & +0.058 [-0.050, +0.167] & -0.008 [-0.067, +0.058] & +0.258 [+0.158, +0.358] & 0.150 \\
Phi-4 & +0.208 [+0.117, +0.300] & +0.050 [+0.000, +0.108] & +0.400 [+0.300, +0.508] & 0.160 \\
Llama-3.1-8B & +0.125 [+0.042, +0.217] & -0.133 [-0.217, -0.050] & +0.050 [-0.058, +0.158] & 0.160 \\
\midrule
\multicolumn{7}{@{}l}{\emph{component accuracies behind the contrasts above}} \\
backbone & rigged at max $k$ & fault at max $k$ & clean at max $k$ & direct solve & fault gain & no-fault gain \\
Qwen3-14B & 0.575 & 0.600 & 0.867 & 0.833 & +0.392 & +0.300 \\
Qwen3-8B & 0.633 & 0.575 & 0.808 & 0.817 & +0.375 & +0.117 \\
Phi-4 & 0.792 & 0.583 & 0.825 & 0.775 & +0.392 & -0.008 \\
Llama-3.1-8B & 0.492 & 0.367 & 0.383 & 0.517 & +0.233 & +0.183 \\
\midrule
\multicolumn{7}{@{}l}{\emph{the interaction at every dose (exploratory: the registration fixes $D$ at max $k$)}} \\
backbone & $D$ at $k{=}1$ & $D$ at $k{=}2$ & $D$ at $k{=}3$ & power at the observed $D$ & $n$ & power target \\
Qwen3-14B & +0.192 [+0.075, +0.308] & +0.092 [-0.025, +0.208] & +0.092 [-0.025, +0.208] & 0.346 & 120 & 0.800 \\
Qwen3-8B & +0.258 [+0.158, +0.367] & +0.258 [+0.158, +0.358] & +0.258 [+0.158, +0.358] & 0.999 & 120 & 0.800 \\
Phi-4 & +0.375 [+0.258, +0.492] & +0.375 [+0.267, +0.483] & +0.400 [+0.300, +0.508] & 1.000 & 120 & 0.800 \\
Llama-3.1-8B & +0.025 [-0.075, +0.125] & +0.017 [-0.092, +0.125] & +0.050 [-0.058, +0.158] & 0.165 & 120 & 0.800 \\
\midrule
\multicolumn{7}{@{}l}{\emph{the length-matched rigging control (exploratory, post-pin): a content-free upstream of the SAME length}} \\
backbone & rig-len acc. & rig-len $-$ rig [95\%] & rig-len $-$ fault [95\%] & McNemar $p$ & status \\
Qwen3-14B & 0.408 & -0.167 [-0.250, -0.083] & -0.192 [-0.308, -0.067] & 3.2e-4 & landed \\
Qwen3-8B & 0.600 & -0.033 [-0.125, +0.058] & +0.025 [-0.083, +0.133] & 0.597 & landed \\
Phi-4 & 0.825 & +0.033 [+0.000, +0.075] & +0.242 [+0.150, +0.333] & 0.219 & landed \\
Llama-3.1-8B & 0.425 & -0.067 [-0.133, +0.000] & +0.058 [-0.033, +0.150] & 0.077 & landed \\
\bottomrule
\end{tabular}
\vspace{0.4em}
\begin{minipage}{\linewidth}\footnotesize
\textbf{Detectable $D$} is the smallest interaction this design detects at the stated power, simulated over the landed per-item vectors. Per item i the interaction contributes d\_i = (F\_k,i - F\_0,i) - (C\_k,i - C\_0,i), a value in \{-2,-1,0,1,2\}; D is the mean of d. The LANDED d vector is centred to mean 0, which fixes the dispersion at the one the campaign actually measured and removes the observed effect. For a candidate true effect delta: draw n items with replacement from the centred vector, add delta, and reject when the 95\% paired-item interval of the resampled mean excludes 0. The interval in the simulation is mean +- 1.96 * sd/sqrt(n), the normal-approximation form of the same paired-item bootstrap, and its half-width is recorded against the analyser's own percentile bootstrap on the SAME landed vector, so the approximation is measured rather than assumed. Power at delta is the rejection rate over the simulated draws; the MDE is the smallest delta on the grid whose power reaches the target.
\end{minipage}
\end{table}

%% file: tables/T11-strict-direct.tex
\begin{table}[t]
\centering
\footnotesize
\caption{The strict pipeline against a single direct call, and dependence on the upstream message. Every block here is exploratory and two-sided: no direction was written down before the data. The first block contrasts the \emph{strict} no-fault pipeline at $k{=}0$ with one direct call at the same answer format; Table~\ref{tab:mechanism} contrasts the fully re-grounded no-fault cell instead, which is a different comparison. The second block repeats it with full message history. The third reports what the rigging controls support: no equivalence margin is registered, so no equivalence test is run, and a non-significant interval is reported as a failure to detect with the effect it still admits. That block carries BOTH rigging arms and says which one each verdict comes from, because the registered one-line control replaces the upstream message with a shorter string as well as an emptier one, and the length-matched arm is the one that separates the two. The blocks are, in order: the strict four-stage pipeline against one call; the same contrast with full message history; dependence on the upstream message: which control decides, and what its interval supports.}
\label{tab:strict-direct}
\begin{tabular}{lp{1.15in}lllp{1.5in}}
\toprule
\multicolumn{6}{@{}l}{\emph{the strict four-stage pipeline against one call}} \\
backbone & strict no-fault & direct solve & difference [95\%] & McNemar $p$ \\
Qwen3-14B & 0.567 & 0.833 & -0.267 [-0.367, -0.167] & 9.4e-7 \\
Qwen3-8B & 0.692 & 0.817 & -0.125 [-0.208, -0.050] & 0.004 \\
Phi-4 & 0.833 & 0.775 & +0.058 [+0.000, +0.125] & 0.118 \\
Llama-3.1-8B & 0.200 & 0.517 & -0.317 [-0.417, -0.217] & 1.4e-8 \\
\midrule
\multicolumn{6}{@{}l}{\emph{the same contrast with full message history}} \\
backbone & strict no-fault, full history & direct solve & difference [95\%] & McNemar $p$ \\
Qwen3-14B & 0.783 & 0.833 & -0.050 [-0.125, +0.025] & 0.263 \\
Qwen3-8B & 0.683 & 0.817 & -0.133 [-0.217, -0.058] & 0.002 \\
Phi-4 & 0.825 & 0.775 & +0.050 [-0.017, +0.117] & 0.210 \\
Llama-3.1-8B & 0.250 & 0.517 & -0.267 [-0.358, -0.175] & 1.9e-7 \\
\midrule
\multicolumn{6}{@{}l}{\emph{dependence on the upstream message: which control decides, and what its interval supports}} \\
backbone & rigging arm & arm $-$ fault at max $k$ & [95\%] & detected? & statement \\
Qwen3-14B & rig\_k3 (one-line, registered) & -0.025 & [-0.142, +0.092] & no & reported, does not decide: shorter as well as emptier \\
 & rig\_k3\_len (length-matched, post-pin) & -0.192 & [-0.308, -0.067] & yes & decides: content at fixed length costs accuracy \\
Qwen3-8B & rig\_k3 (one-line, registered) & +0.058 & [-0.050, +0.167] & no & reported, does not decide: shorter as well as emptier \\
 & rig\_k3\_len (length-matched, post-pin) & +0.025 & [-0.083, +0.133] & no & decides: none detected; the interval admits $\pm$0.108 \\
Phi-4 & rig\_k3 (one-line, registered) & +0.208 & [+0.117, +0.300] & yes & reported, does not decide: shorter as well as emptier \\
 & rig\_k3\_len (length-matched, post-pin) & +0.242 & [+0.150, +0.333] & yes & decides: content-free upstream scores \emph{above} the faulty arm \\
Llama-3.1-8B & rig\_k3 (one-line, registered) & +0.125 & [+0.042, +0.217] & yes & reported, does not decide: shorter as well as emptier \\
 & rig\_k3\_len (length-matched, post-pin) & +0.058 & [-0.033, +0.150] & no & decides: none detected; the interval admits $\pm$0.092 \\
\bottomrule
\end{tabular}
\vspace{0.4em}
\begin{minipage}{\linewidth}\footnotesize
No equivalence test is reported. No equivalence margin is registered anywhere in the analysis plan, so no TOST is run: a TOST needs a margin declared before the data, and choosing one now would set the threshold after seeing the interval it has to clear.
\\ Decomposition, (rig\_k3\_len $-$ fault) $=$ (rig\_k3\_len $-$ rig\_k3) $+$ (rig\_k3 $-$ fault): \textbf{Qwen3-14B}: -0.192 $=$ -0.167 $+$ -0.025; \textbf{Qwen3-8B}: +0.025 $=$ -0.033 $+$ +0.058; \textbf{Phi-4}: +0.242 $=$ +0.033 $+$ +0.208; \textbf{Llama-3.1-8B}: +0.058 $=$ -0.067 $+$ +0.125.
\end{minipage}
\end{table}

%% file: tables/T13-defense-n120.tex
\begin{table}[t]
\centering
\footnotesize
\caption{The defence head-to-head at the headline standard: every arm on the same items, with per-arm intervals and paired contrasts. The first block is arm accuracy with $95\%$ Wilson intervals. The second is the two contrasts the comparison turns on and TWO ceiling ratios: the matched one divides the grounded inspector by the isomorphic no-fault cell at the same stage count, the legacy one by a no-fault cell at a different stage count. Both are printed because they are different quantities. Exploratory and two-sided throughout. The blocks are, in order: arm accuracy, same items, 95\% Wilson intervals: no-fault arms; arm accuracy, same items, 95\% Wilson intervals: faulted arms; the two contrasts, and the two ceiling ratios; what an ungrounded inspector costs a pipeline with NO fault (clean\_inspect $-$ clean, same items).}
\label{tab:defense}
\begin{tabular}{lrrrrl}
\toprule
\multicolumn{6}{@{}l}{\emph{arm accuracy, same items, 95\% Wilson intervals: no-fault arms}} \\
backbone & clean (defence file) & clean+insp & clean+insp k4 &  &  \\
Qwen3-14B & 0.567 [0.477, 0.652] & 0.567 [0.477, 0.652] & 0.833 [0.757, 0.889] &  &  \\
Qwen3-8B & 0.667 [0.578, 0.745] & 0.692 [0.604, 0.767] & 0.792 [0.711, 0.855] &  &  \\
Phi-4 & 0.825 [0.747, 0.883] & 0.633 [0.544, 0.714] & 0.808 [0.729, 0.869] &  &  \\
Llama-3.1-8B & 0.200 [0.138, 0.280] & 0.067 [0.034, 0.126] & 0.433 [0.348, 0.523] &  &  \\
\midrule
\multicolumn{6}{@{}l}{\emph{arm accuracy, same items, 95\% Wilson intervals: faulted arms}} \\
backbone & strict & blind insp & grounded insp &  &  \\
Qwen3-14B & 0.208 [0.145, 0.289] & 0.208 [0.145, 0.289] & 0.817 [0.738, 0.876] &  &  \\
Qwen3-8B & 0.200 [0.138, 0.280] & 0.200 [0.138, 0.280] & 0.775 [0.692, 0.841] &  &  \\
Phi-4 & 0.183 [0.124, 0.262] & 0.175 [0.117, 0.253] & 0.783 [0.701, 0.848] &  &  \\
Llama-3.1-8B & 0.133 [0.084, 0.206] & 0.058 [0.029, 0.116] & 0.417 [0.332, 0.506] &  &  \\
\midrule
\multicolumn{6}{@{}l}{\emph{the two contrasts, and the two ceiling ratios}} \\
backbone & grounded $-$ blind [95\%] & grounded $-$ strict [95\%] & grounded / clean+insp k4 (matched) & grounded / clean+insp (legacy) & $n$ \\
Qwen3-14B & +0.608 [+0.517, +0.700] & +0.608 [+0.517, +0.700] & 0.980 [0.931, 1.030] & 1.441 [1.247, 1.695] & 120 \\
Qwen3-8B & +0.575 [+0.483, +0.667] & +0.575 [+0.483, +0.667] & 0.979 [0.928, 1.031] & 1.120 [1.000, 1.257] & 120 \\
Phi-4 & +0.608 [+0.500, +0.708] & +0.600 [+0.492, +0.700] & 0.969 [0.929, 1.000] & 1.237 [1.083, 1.431] & 120 \\
Llama-3.1-8B & +0.358 [+0.267, +0.458] & +0.283 [+0.183, +0.383] & 0.962 [0.800, 1.150] & 6.250 [3.500, 16.667] & 120 \\
\midrule
\multicolumn{6}{@{}l}{\emph{what an ungrounded inspector costs a pipeline with NO fault (clean\_inspect $-$ clean, same items)}} \\
backbone & no-fault & $+$ blind inspector & cost [95\%] & McNemar $p$ & items \\
Qwen3-14B & 0.567 & 0.567 & +0.000 [+0.000, +0.000] & 1.000 & +0.0 \\
Qwen3-8B & 0.667 & 0.692 & +0.025 [+0.000, +0.058] & 0.250 & +3.0 \\
Phi-4 & 0.825 & 0.633 & -0.192 [-0.267, -0.117] & 1.5e-6 & -23.0 \\
Llama-3.1-8B & 0.200 & 0.067 & -0.133 [-0.200, -0.075] & 3.1e-5 & -16.0 \\
\bottomrule
\end{tabular}
\vspace{0.4em}
\begin{minipage}{\linewidth}\footnotesize
\emph{clean (defence file)} is the no-fault arm of the defence run, not the no-fault arm of the dose run that Table~\ref{tab:headline} reports at $k{=}0$. Same nominal arm, two separate measurements, so they are not expected to agree and neither corrects the other. They differ on 3 items on Qwen3-8B and 1 item on Phi-4; elsewhere they agree item for item.
The matched ratio's denominator is the inspector pipeline with no fault at the same stage count; the legacy one's is a no-fault cell at a different stage count. A ratio above 1 under the legacy denominator and below it under the matched one is a statement about the denominator, not about the inspector.
\\ The last block is the cost of the inspector where there is nothing to inspect: the same no-fault pipeline with and without a blind reviewer stage, on the same items. It is not a defence contrast and is not comparable with the rows above, which all start from the faulty pipeline.
\end{minipage}
\end{table}

%% file: tables/T8-pipelines.tex
\begin{table}[t]
\centering
\footnotesize
\caption{The two measured pipelines side by side. Pipeline~A is the registered four-stage pipeline; pipeline~B is a three-stage pipeline measured on the same items, with its own matched ceilings, and is \emph{exploratory}: it is not in the registered analysis plan and enters no confirmatory family. Each pipeline's interaction is taken at its OWN max $k$, which differs between them and is printed. $D$ is the fault by re-grounding interaction with a paired item bootstrap interval; a $D$ whose interval excludes 0 is the fault-specific part of the re-grounding gain.}
\label{tab:pipelines}
\begin{tabular}{llrrllrrl}
\toprule
 &  &  &  & \multicolumn{2}{c}{PRR, min to max $k$} & \multicolumn{2}{c}{gain} &  \\
\cmidrule(lr){5-6}\cmidrule(lr){7-8}
backbone & pipeline & stages & max $k$ & matched & deployed & fault & no-fault & $D$ at max $k$ [95\%] \\
\midrule
Qwen3-14B & A (registered) & 4 & 3 & 0.368\,$\to$\,0.692 & 0.368\,$\to$\,1.059 & +0.392 & +0.300 & +0.092 [-0.025, +0.208] \\
 & B (exploratory) & 3 & 2 & 0.137\,$\to$\,0.647 & 0.137\,$\to$\,0.647 & +0.433 & +0.000 & +0.433 [+0.342, +0.525] \\
Qwen3-8B & A (registered) & 4 & 3 & 0.289\,$\to$\,0.711 & 0.289\,$\to$\,0.831 & +0.375 & +0.117 & +0.258 [+0.158, +0.358] \\
 & B (exploratory) & 3 & 2 & 0.040\,$\to$\,0.293 & 0.040\,$\to$\,0.287 & +0.208 & -0.017 & +0.225 [+0.142, +0.308] \\
Phi-4 & A (registered) & 4 & 3 & 0.230\,$\to$\,0.707 & 0.230\,$\to$\,0.700 & +0.392 & -0.008 & +0.400 [+0.300, +0.508] \\
 & B (exploratory) & 3 & 2 & 0.196\,$\to$\,0.531 & 0.196\,$\to$\,0.526 & +0.267 & -0.008 & +0.275 [+0.192, +0.358] \\
Llama-3.1-8B & A (registered) & 4 & 3 & 0.667\,$\to$\,0.957 & 0.667\,$\to$\,1.833 & +0.233 & +0.183 & +0.050 [-0.058, +0.158] \\
 & B (exploratory) & 3 & 2 & 0.232\,$\to$\,0.593 & 0.232\,$\to$\,0.625 & +0.183 & +0.025 & +0.158 [+0.042, +0.275] \\
\midrule
\multicolumn{9}{@{}l}{\emph{Summary, on the same items:} $D$ excludes 0 on 2 of 4 backbones under pipeline~A and on 4 of 4 under pipeline~B.} \\
\bottomrule
\end{tabular}
\vspace{0.4em}
\begin{minipage}{\linewidth}\footnotesize
Intervals are $95\%$ paired item bootstraps. Pipeline~B is reported because the registration's exclusion rule is \emph{none post hoc}: it is a measured design arm, so it is either reported or its exclusion is dated. It is labelled exploratory everywhere because it is not in the registered plan, and no kill clause and no confirmatory family reads it.
\end{minipage}
\end{table}

%% file: tables/T12-history-full.tex
\begin{table}[t]
\centering
\footnotesize
\caption{External validity: the same pipeline with full message history against the registered predecessor-only setting. The registered arms pass each stage only its predecessor's message; production frameworks pass the whole history, so this condition asks how much of the dose response is a property of re-grounding rather than of the restricted setting. Exploratory: the full-history arm is not in the registered plan, and the contrast is two-sided because its direction was not written down before the data.}
\label{tab:history}
\begin{tabular}{lllllll}
\toprule
 & \multicolumn{2}{c}{PRR, min to max $k$} &  &  &  \\
\cmidrule(lr){2-3}
backbone & matched & deployed & $D$ at max $k$ [95\%] & full $-$ predecessor at max $k$ [95\%] & McNemar $p$ \\
\midrule
Qwen3-14B & 0.277\,$\to$\,0.706 & 0.277\,$\to$\,0.766 & +0.317 [+0.200, +0.433] & +0.000 [-0.042, +0.042] & 1.000 \\
Qwen3-8B & 0.256\,$\to$\,0.700 & 0.256\,$\to$\,0.854 & +0.258 [+0.150, +0.359] & +0.008 [-0.033, +0.050] & 1.000 \\
Phi-4 & 0.212\,$\to$\,0.660 & 0.212\,$\to$\,0.667 & +0.367 [+0.267, +0.467] & -0.033 [-0.075, +0.000] & 0.219 \\
Llama-3.1-8B & 0.267\,$\to$\,0.660 & 0.267\,$\to$\,1.167 & +0.033 [-0.083, +0.150] & -0.075 [-0.133, -0.017] & 0.022 \\
\bottomrule
\end{tabular}
\vspace{0.4em}
\begin{minipage}{\linewidth}\footnotesize
Intervals are $95\%$ paired item bootstraps on the same items. The contrast column is the full-history arm minus the registered predecessor-only arm at max $k$, paired item by item.
\end{minipage}
\end{table}

%% file: tables/T9-cost-headline.tex
\begin{table}[t]
\centering
\footnotesize
\caption{The cost of re-grounding at the headline standard. Every row is $n$ items per arm on the pipeline task, on four backbones. Average tokens are prompt \emph{and} completion per item, the same quantity the cost figure's axis carries. Both registered ratios are shown because they answer different questions: PRR (matched) divides by the moving matched ceiling: the no-fault cell at the same $k$ (\texttt{clean\_k} at that $k$), PRR (deployed) by the deployed strict pipeline at $k{=}0$. ONE ROUNDING RULE governs the two derived columns: the token endpoints are rounded to the printed figure first and then subtracted, so $\Delta$tokens is exactly the difference of the two token values printed above and below it, and the marginal column is exactly the printed ratio step divided by that printed $\Delta$.}
\label{tab:cost-headline}
\begin{tabular}{llrrlrrr}
\toprule
backbone & $k$ & acc. & avg.\ prompt$+$completion & PRR (matched) [95\%] & PRR (deployed) & $\Delta$tokens & marginal \\
\midrule
Qwen3-14B & 0 & 0.208 & 917.5 & 0.368 [0.243, 0.508] & 0.368 & -- & -- \\
 & 1 & 0.533 & 977.3 & 0.762 [0.643, 0.888] & 0.941 & +59.8 & +6.6 \\
 & 2 & 0.600 & 1065.7 & 0.692 [0.598, 0.781] & 1.059 & +88.4 & -0.8 \\
 & 3 & 0.600 & 1133.4 & 0.692 [0.598, 0.781] & 1.059 & +67.7 & +0.0 \\
 & $n$ & 120 & items per arm &  &  &  &  \\
\midrule
Qwen3-8B & 0 & 0.200 & 1009.4 & 0.289 [0.189, 0.397] & 0.289 & -- & -- \\
 & 1 & 0.558 & 1040.6 & 0.705 [0.596, 0.812] & 0.807 & +31.2 & +13.3 \\
 & 2 & 0.575 & 1148.8 & 0.711 [0.606, 0.814] & 0.831 & +108.2 & +0.1 \\
 & 3 & 0.575 & 1215.9 & 0.711 [0.606, 0.814] & 0.831 & +67.1 & +0.0 \\
 & $n$ & 120 & items per arm &  &  &  &  \\
\midrule
Phi-4 & 0 & 0.192 & 1186.7 & 0.230 [0.150, 0.317] & 0.230 & -- & -- \\
 & 1 & 0.533 & 1201.3 & 0.667 [0.560, 0.775] & 0.640 & +14.6 & +29.9 \\
 & 2 & 0.575 & 1214.5 & 0.683 [0.583, 0.782] & 0.690 & +13.2 & +1.2 \\
 & 3 & 0.583 & 1276.8 & 0.707 [0.608, 0.806] & 0.700 & +62.3 & +0.4 \\
 & $n$ & 120 & items per arm &  &  &  &  \\
\midrule
Llama-3.1-8B & 0 & 0.133 & 753.5 & 0.667 [0.407, 1.000] & 0.667 & -- & -- \\
 & 1 & 0.258 & 825.8 & 0.861 [0.639, 1.139] & 1.292 & +72.3 & +2.7 \\
 & 2 & 0.333 & 874.1 & 0.870 [0.680, 1.098] & 1.667 & +48.3 & +0.2 \\
 & 3 & 0.367 & 983.9 & 0.957 [0.760, 1.204] & 1.833 & +109.8 & +0.8 \\
 & $n$ & 120 & items per arm &  &  &  &  \\
\bottomrule
\end{tabular}
\vspace{0.4em}
\begin{minipage}{\linewidth}\footnotesize
Intervals are $95\%$ paired item bootstraps on the ratio. Token definition: avg\_tokens(arm) = (prompt\_tokens|arm + completion\_tokens|arm) / n, read from the recorded \texttt{usage} of each arm.
\end{minipage}
\end{table}

%% file: tables/T6-contribution-three.tex
\begin{table}[t]
\centering
\footnotesize
\caption{Contribution 3, robustness-aware topology selection, reported as a negative control. The prospective search picks on a development split and is scored on held-out topologies; the retrospective front is computed over the whole bank. The retrospective advantage compares fronts that differ in cost and in clean accuracy and is therefore unmatched; the cost-tier gain is the matched contrast. The blocks are, in order: prospective search, scored on the held-out split; retrospective front over the whole topology bank.}
\label{tab:contribution-three}
\begin{tabular}{llrrrp{1.9in}}
\toprule
\multicolumn{6}{@{}l}{\emph{prospective search, scored on the held-out split}} \\
block & arm & picks & clean acc. & held-out PRR & note \\
search & robustness-aware (ours) & 3 & 0.850 & 0.985 & best-in-front PRR 0.985 \\
 & accuracy-only & 3 & 0.842 & 0.990 & best-in-front PRR 1.000 \\
 & random & 5 & 0.855 & 0.991 & best-in-front PRR 1.000 \\
contrast & ours $-$ accuracy-only &  &  & -0.5 pts & matched: same held-out split, same scoring \\
contrast & ours $-$ random &  &  & -0.6 pts & unmatched: fronts of 3 and 5 \\
contrast & accuracy-only $-$ random &  &  & -0.1 pts & unmatched: fronts of 3 and 5 \\
interval & none at this $n$ &  &  & -- & the search result file stores per-arm means only; an interval needs the per-topology values and they are not recorded \\
\midrule
\multicolumn{6}{@{}l}{\emph{retrospective front over the whole topology bank}} \\
block & front & size & clean acc. & front PRR & note \\
front & robustness-aware front & 2 & 0.830 & 0.956 & front cost 12 \\
 & accuracy-only front & 6 & 0.823 & 0.949 & front cost 8 \\
contrast & advantage (unmatched) &  &  & +0.7 pts & the fronts differ in cost and in clean accuracy \\
contrast & cost-tier gain (matched) &  & at cost 6 & +0.0 pts & both strategies pick the same topology at the tier \\
bank & records scored & 30 & candidates at the tier & 10 & the bank both fronts are drawn from \\
\bottomrule
\end{tabular}
\vspace{0.4em}
\begin{minipage}{\linewidth}\footnotesize
The two blocks say the same thing: the matched contrast is 0 retrospectively and negative prospectively. Contribution 3 stays in the paper as a negative control, not as a claim.
\end{minipage}
\end{table}

%% file: tables/T5-flat-regime.tex
\begin{table}[t]
\centering
\footnotesize
\caption{The flat regime on MATH500: the structural predictor on the topology bank, the operator against a one-line proximity heuristic, and per-topology subset ranking. Single backbone, exploratory, reported as measured. A correlation whose interval covers zero is not evidence of a predictor. Each block names its own columns, because the four blocks report different quantities and one shared header would mislabel three of them. The blocks are, in order: FIM sink-influence against measured PRR, per fault family (MATH500 topology bank); registered standard benchmarks for the flat regime; tie battery: operator against the one-line heuristic; subset ranking, per topology (Spearman $\rho$ of predicted danger against measured PRR).}
\label{tab:flat}
\begin{tabular}{lllllp{1.6in}}
\toprule
\multicolumn{6}{@{}l}{\emph{FIM sink-influence against measured PRR, per fault family (MATH500 topology bank)}} \\
quantity & fault family & $\rho$ [95\%] & $|\rho|$ & $n$ & verdict \\
Spearman correlation & reasoning error & -0.503 [-0.756, -0.166] & 0.503 & 30 & informative \\
 & malicious noise & -0.024 [-0.442, +0.390] & 0.024 & 30 & covers zero \\
 & role drift & -0.055 [-0.466, +0.370] & 0.055 & 30 & covers zero \\
bank & topologies scored & 30 & items each & 50 & the bank the rows above are computed on \\
\midrule
\multicolumn{6}{@{}l}{\emph{registered standard benchmarks for the flat regime}} \\
quantity & benchmark & topologies measured &  &  & status \\
coverage & MATH500 & 30 &  &  & landed; the bank the rows above use \\
 & MMLU & 30 &  &  & landed; the full bank \\
 & gsm\_hard & -- &  &  & the pipeline task; no bank of its own \\
\midrule
\multicolumn{6}{@{}l}{\emph{tie battery: operator against the one-line heuristic}} \\
contrast & cell & operator $|\rho|$ & heuristic $|\rho|$ & $\Delta$ [95\%] & $n$, flag \\
operator $-$ heuristic & A / reasoning error & 0.503 & 0.492 & +0.011 [-0.302, +0.312] & 30, ties \\
 & A / wrong intermediate & not run & not run & not run & 0; the bank injects reasoning error, malicious noise, role drift \\
 & B / pairs & 0.198 & 0.242 & -0.044 [-0.200, +0.122] & 90, ties \\
multifault & C / convex against structural & 0.678 & 0.740 & -0.062 & 8 topologies, no interval \\
multifault bank & topologies carried &  &  & 8 & the bank block C consumes \\
\midrule
\multicolumn{6}{@{}l}{\emph{subset ranking, per topology (Spearman $\rho$ of predicted danger against measured PRR)}} \\
quantity & topology & $\rho$ FIM & $\rho$ structural &  & better predictor \\
subset ranking & gen022 & -0.972 & -0.972 &  & tie \\
 & gen038 & -0.846 & -0.893 &  & structural \\
 & gen070 & -0.886 & -0.886 &  & tie \\
 & gen074 & -0.229 & -0.589 &  & structural \\
 & gen018 & -0.874 & -0.679 &  & FIM \\
 & gen034 & -0.267 & -0.564 &  & structural \\
 & gen020 & -0.648 & -0.797 &  & structural \\
 & gen007 & -0.749 & -0.679 &  & FIM \\
\bottomrule
\end{tabular}
\vspace{0.4em}
\begin{minipage}{\linewidth}\footnotesize
Every block here is exploratory and sits outside the corrected families. The tie battery's own file records that its prose anchor was never a measured value; the measured operator-minus-heuristic contrast is the row above, and it is the one the paper reports.
\end{minipage}
\end{table}